\documentclass[10pt,twocolumn,letterpaper]{article}

\usepackage{iccv}
\usepackage{times}
\usepackage{epsfig}
\usepackage{graphicx}
\usepackage{amsmath}
\usepackage{amssymb}

\usepackage{color}
\usepackage[dvipsnames]{xcolor}
\usepackage{subcaption}
\usepackage[accsupp]{axessibility}  % Improves PDF readability for those with disabilities.
\usepackage{pifont} % Access to PostScript standard Symbol and Dingbats fonts
\usepackage{dsfont} % Math symbols
\usepackage{multirow} 
\usepackage{booktabs} 

\usepackage{algorithm}
\usepackage{algpseudocode} 
\usepackage{listings}

\newcommand{\authorsep}{\hspace{8pt}}
\usepackage[breaklinks=true,bookmarks=false]{hyperref}

\iccvfinalcopy % *** Uncomment this line for the final submission

\def\iccvPaperID{****} % *** Enter the ICCV Paper ID here

\ificcvfinal\fi

\begin{document}

%%%%%%%%% TITLE
\title{A Unified Continual Learning Framework with General \\
  Parameter-Efficient Tuning }

\author{
  Qiankun Gao$^{1}$\authorsep Chen Zhao$^\dagger$$^{2}$\authorsep Yifan Sun$^{3}$\authorsep Teng Xi$^{3}$\authorsep Gang Zhang$^{3}$\authorsep Bernard Ghanem$^{2}$\authorsep Jian Zhang$^\dagger$$^{1}$ \\
  \small$^1$Peking University Shenzhen Graduate School \hspace{5pt}
  \small$^2$King Abdullah University of Science and Technology (KAUST) \hspace{5pt}\small$^3$Baidu Inc.\\
  \small gqk@stu.pku.edu.cn \hspace{5pt}
  \small chen.zhao@kaust.edu.sa \hspace{5pt}
  \small zhangjian.sz@pku.edu.cn\\
}

\maketitle
\let\thefootnote\relax\footnotetext{$^\dagger$ Corresponding authors.}

%%%%%%%%% ABSTRACT
\begin{abstract}
  The ``pre-training → downstream adaptation'' presents both new opportunities and challenges for Continual Learning (CL). Although the recent state-of-the-art in CL is achieved through Parameter-Efficient-Tuning (PET) adaptation paradigm, only prompt has been explored, limiting its application to Transformers only. In this paper, we position prompting as one instantiation of PET, and propose a unified CL framework with general PET, dubbed as Learning-Accumulation-Ensemble (LAE). PET, \eg, using Adapter, LoRA, or Prefix, can adapt a pre-trained model to downstream tasks with fewer parameters and resources. Given a PET method, our LAE framework incorporates it for CL with three novel designs. 1) Learning: the pre-trained model adapts to the new task by tuning an online PET module, along with our adaptation speed calibration to align different PET modules, 2) Accumulation: the task-specific knowledge learned by the online PET module is accumulated into an offline PET module through momentum update, 3) Ensemble: During inference, we respectively construct two experts with online/offline PET modules (which are favored by the novel/historical tasks) for prediction ensemble. We show that LAE is compatible with a battery of PET methods and gains strong CL capability. For example, LAE with Adaptor PET surpasses the prior state-of-the-art by 1.3\% and 3.6\% in last-incremental accuracy on CIFAR100 and ImageNet-R datasets, respectively. Code is available at \url{https://github.com/gqk/LAE}.
\end{abstract}

%%%%%%%%% BODY TEXT
\vspace{-15pt}
\section{Introduction}
\label{sec.introduction}
Continual Learning (CL) of new knowledge is an essential ability for AI models in the constantly changing world. However, neural networks often suffer from catastrophic forgetting~\cite{catastrophic_forgetting,cfrp}, in which previously learned knowledge is forgotten when the model incorporates novel information. Although many works have been devoted to reducing forgetting, such as dynamic networks~\cite{pnn,den,l2g,cpg}, regularization~\cite{lwf,ewc,pi,mas}, and memory replay~\cite{icarl,ucir,podnet,aanets,alssum2023smile}, their performance still falls short of practical requirements.

\begin{figure}[t]
  \begin{center}
    \includegraphics[width=\linewidth]{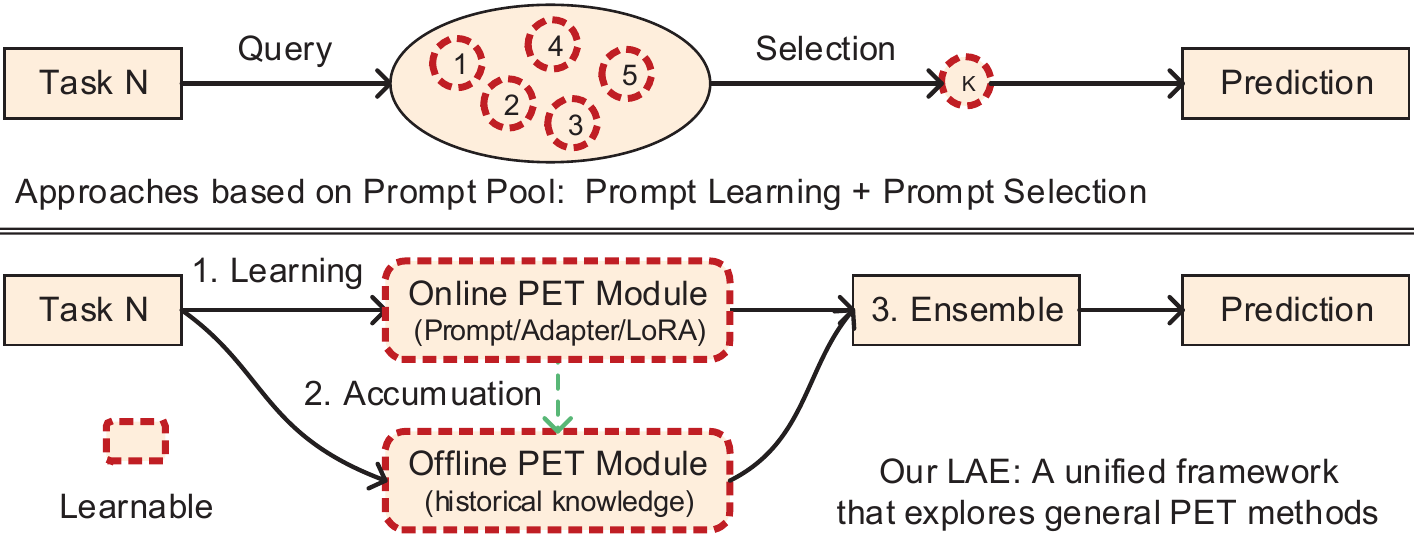}
  \end{center}
  \caption{\textbf{The pipeline of our LAE framework \vs prompt-pool approaches.} \textbf{Above:} Prompt-pool approaches, which query prompts from a pool of learnable prompts. \textbf{Below:} Proposed LAE, where  \textit{an online Parameter-Efficient Tuning (PET) module} attached to the pre-trained model to adapt a new task quickly,  and \textit{an offline PET module} accumulates the learned knowledge slowly. During inference, we use the ensemble of the predictions of the online and offline PET modules as the final prediction.
  }
  \vskip -0.1in
  \label{fig:overview}
\end{figure}

Recently, pre-training and downstream adaptation techniques have opened up new opportunities and challenges for CL. Basically, these techniques ~\cite{bert, gpt, moco, mae, ufo} pre-train a deep model on large-scale data and then adapt the pre-trained model to novel tasks.
We observe that downstream adaptation and CL are important for each other.
On the one hand, in realistic AI systems, pre-trained models sometimes needs to be adapted to multiple downstream tasks sequentially, yielding the need of CL. On the other hand, recent efforts~\cite{l2p, dual_prompt,esn} show that the ``pre-training $\rightarrow$ downstream adaptation'' techniques can boost CL performance.

Specifically, L2P~\cite{l2p}, DualPrompt~\cite{ dual_prompt}, and ESN~\cite{esn} all use a popular adaptation technique named Parameter-Efficient-Tuning (PET).
Generally, PET adapts pre-trained models to downstream tasks with much fewer learnable parameters, as well as fewer resources.
Though these approaches have advanced the state-of-the-art in CL, they still have some limitations. 1) They are all constrained to a specific PET method, \emph{i.e.}, prompt tuning,
limiting their flexibility, considering that prompt can only cooperate with transformers and does not accommodate other network architectures. 2) Most of them rely on selecting task-specific parameters (the prompt tokens, in particular) for each individual task. The selection tends to be noisy with increasing task numbers and the task-specific prompts appear homogeneous, according to our investigation in the supplementary.

\begin{figure*}[ht]
  \begin{center}
    \centerline{\includegraphics[width=\textwidth]{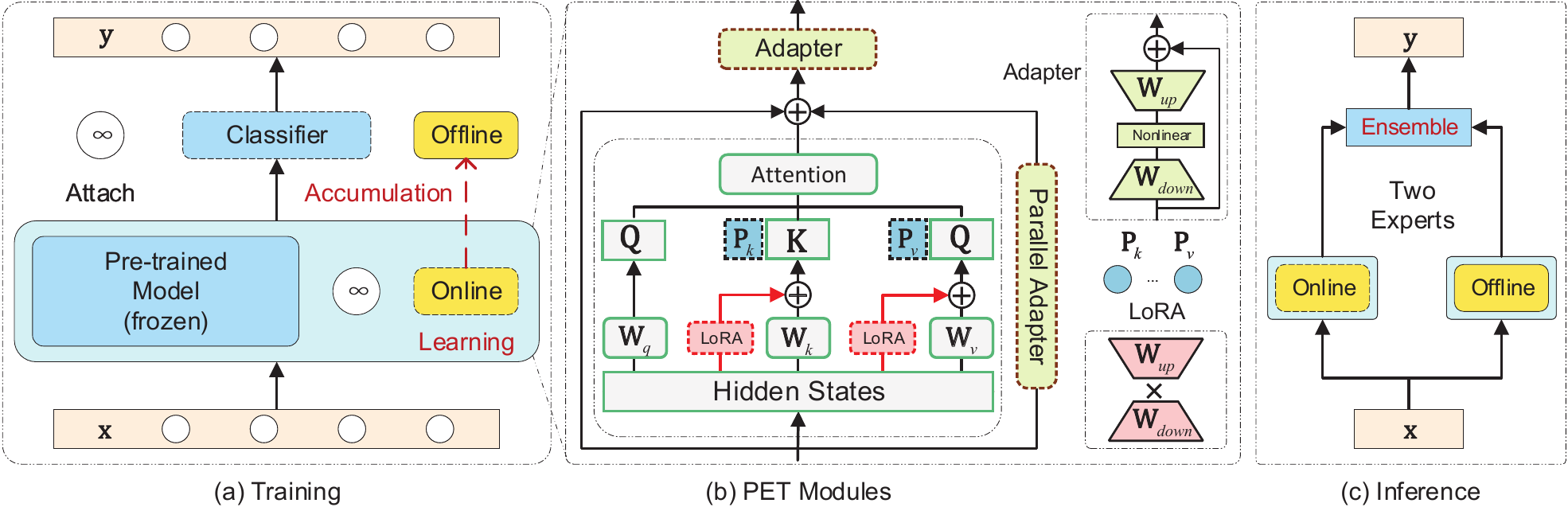}}
    \caption{\textbf{Illustration of our LAE framework.} The left (a) is the training process, the right (c) is the inference flow, and the middle (b) lists some representative Parameter-Efficient Tuning (PET) modules attached to a transformer attention block, modules connected with dashed lines are optional and we use one of the PET modules in the experiments. There is no residual connection in Parallel Adapter. ``Online'' is a PET module to learn knowledge from the new task and ``Offline'' is a PET module to accumulate knowledge. The pre-trained model is omitted in the inference flow for concise.
    }
    \label{fig:framework}
  \end{center}
  \vskip -0.4in
\end{figure*}

To circumvent these issues, this paper proposes Learning-Accumulation-Ensemble (LAE), a unified CL framework resort to the general Parameter-Efficient Tuning (PET). LAE is not restricted to Prompt, but can also utilize various other PET modules as shown in Fig.~\ref{fig:framework} (b).
Given a PET method, our LAE directly reshapes it for CL with three steps, \emph{i.e.}, learning, accumulation, and ensemble.

$\bullet$ \textbf{1) Learning with calibrated speed.}
The pre-trained model adapts to the new task by tuning an online PET module. To accommodate various PET methods, a key challenge is that different PET modules have different \textit{adaptation speeds} (for novel tasks), as well as different \textit{forgetting speeds} (for historical tasks). In response, we design an adaptation calibration strategy, based on the gradient analysis for different PET modules. We empirically show that this calibration strategy aligns different PET against each other and is critical for LAE to be a unified framework.

$\bullet$ \textbf{2) Accumulation of multi-task knowledge.} After adapting the pre-trained model to a new task,
the parameters in the online PET module are prone to the current novel task and may not fit historical tasks. Instead of memorizing multiple sets of PET modules and selecting some subsets (as in L2P and DualPrompt) for individual tasks, LAE accumulates all the knowledge of already-seen tasks into a single offline PET module through momentum update. This simple accumulation avoids noisy selection and is competent for \textit{alleviating catastrophic forgetting, especially when the amount of learned tasks is large} (Figs.~\ref{fig:benchmarks} and~\ref{fig:ens_ada} in Sec.~\ref{sec:experiment}).

$\bullet$ \textbf{3) Ensemble of two expert models.}
The online and offline PET modules respectively contain more novel and historical knowledge, therefore, two expert models constructed with them are correspondingly better at handling newer and older tasks. Instead of inference only using the online or offline expert model, we integrate the outputs of two expert models by an energy indicator (detailed in Sec.~\ref{sec:framework}) to obtain the prediction for an inference sample from any learned task. This expert ensemble strategy helps our framework to achieve a more robust performance compared to inference using one of the expert models alone.

\smallskip
\noindent The contributions of this paper are summarized as follows:
\begin{itemize}
  \item We thoroughly investigate the novel Continual Learning paradigm that constantly adapts a pre-trained model to novel tasks using general Parameter-Efficient Tuning (PET) methods, and propose a unified Learning-Accumulation-Ensemble (LAE) framework.
  \vskip -0.5in
  \item Our LAE framework reshapes a given PET method into a competitive Memory-Free Continual Learning approach with three novel designs: Learning with calibrated speed, Accumulation of multi-task knowledge, and Ensemble of two expert models constructed with online and offline PET modules.
  \vskip -0.5in
  \item We conduct extensive experiments on CIFAR100 and ImageNet-R benchmarks, on all of which, our LAE consistently achieves superior incremental performance than previous state-of-the-art approaches.
\end{itemize}
\section{Related Works}
\label{sec:rws}

\smallskip
\noindent\textbf{Parameter-Efficient Tuning.} As an efficient alternative to full fine-tuning, Adapter-Tuning~\cite{adapter_tuning} was first proposed to transfer large pre-trained Language models to downstream tasks. Inspired by textual prompting, Prompt-Tuning~\cite{prompt_tuning} and Prefix-Tuning~\cite{prefix_tuning} insert learnable tokens to adapt to the new task. More advanced methods~\cite{bitfit,adaptbias,lora,compacter} achieved comparable or superior performance to full fine-tuning, and keep the same inference cost by merging the additional learnable parameters to the original pre-trained model. Following the step of successful Vision Transformers~\cite{vit,swin}, VPT~\cite{vpt} and AdapterFormer~\cite{adapter_former} have been proposed to solve visual transfer learning problems. Prompt-Tuning and Prefix-Tuning depend on the transformer architecture because they modify input or hidden tokens. Adapter and its variants are network architecture generalizable since they are new modules that can be implemented in forms compatible with pre-trained models. All types of Parameter-Efficient Tuning modules can be integrated into our LAE framework as long as they are suitable for the pre-trained model, but we focus on the representative Adapter~\cite{adapter_tuning}, LoRA~\cite{lora}, and Prefix~\cite{prefix_tuning} in this paper.

\noindent\textbf{Continual Learning.} The central problem of Continual Learning (CL) is fighting catastrophic forgetting~\cite{catastrophic_forgetting}. Memory-based approaches~\cite{icarl,ucir,podnet,aanets,alssum2023smile} save a subset of learned samples into a memory buffer and replay them when learning a new task. Memory-Free approaches do not rely on old samples that may raise privacy concerns, they dynamically expand the network or isolate parameters for different tasks~\cite{pnn,den,l2g,cpg}, regularize the network parameters that are important to learned tasks~\cite{lwf,ewc,pi,mas}, and replay generative or synthetic data~\cite{dgr,deepinversion,abd,r-dfcil}. Conventional CL approaches learn tasks from scratch using a randomly initialized model, while pre-trained models have received little attention from CL researchers until recently. Two pioneering works~\cite{l2p,dual_prompt} introduce Prompt-Tuning to CL and achieve much higher incremental performance than previous approaches, demonstrating the advantage of using pre-trained models in CL. Side-Tuning~\cite{side_tuning} adopts a technique similar to Adapter-Tuning but requires the task identity of the inference sample. In this paper, we propose a unified framework for Memory-Free CL that can incorporate various types of PET modules.
Particularly, we focus on practical Class-Incremental Learning with the potential to extend our LAE to other CL scenarios in future work.

\section{Preliminaries}
\label{sec:preliminaries}

\subsection{Continual Learning Formulation}
We focus on Continual Learning with incremental classes, \ie, Class-Incremental Learning (CIL),
where a model sequentially learns tasks $\mathcal{T}$$:=$$\{\mathcal{T}_1, \mathcal{T}_2, \cdots, \mathcal{T}_n\}$, the $i^{th}$ task $\mathcal{T}_i$ has $|\mathcal{T}_i|$ categories, the train set of $\mathcal{T}_i$ is denoted as $\mathcal{D}_{i}$, and the categories are non-overlapping between tasks. The model $f(\cdot;\boldsymbol{\theta}, \boldsymbol{\phi})$ predicts the category label $y \in \mathcal{Y}$ for a given sample $\mathbf{x} \in \mathcal{X}$ of the learned tasks, where $\mathcal{Y}$ is all seen categories, $\boldsymbol{\theta}$ and $\boldsymbol{\phi}$ are parameters of feature extractor and classification head, respectively. In this paper, the feature extractor is a pre-trained model parameterized by $\boldsymbol{\theta}_{pre}$ attached with the Parameter-Efficient Tuning module parameterized by $\boldsymbol{\theta}_{pet}$, and $\boldsymbol{\phi}$$=$$\operatorname{concatenate}(\boldsymbol{\phi}_{old}, \boldsymbol{\phi}_{new})$, where $\boldsymbol{\phi}_{old}$ and $\boldsymbol{\phi}_{new}$ are classifiers of all learned tasks $\mathcal{T}_{1:i}$ and the current learning task $\mathcal{T}_i$. Since $\boldsymbol{\theta}_{pre}$ and $\boldsymbol{\phi}_{old}$ are kept fixed during learning a new task, we may omit them for concise in the rest of this paper.

\subsection{Parameter-Efficient Tuning Revisit}
Parameter-Efficient Tuning (PET) keeps the pre-trained model frozen and tunes a small number of additional learnable parameters, called PET module in the paper. Below we revisit several representative PET modules, in which $g$ is the module that PET attached to, $\mathbf{e}$ and $\mathbf{h}$ are input and output of the orginal $g$ and $\mathbf{h}'$ is output of $g$ attached with PET.

\smallskip
\noindent\textbf{Adapter}~\cite{adapter_tuning} is a small module that can be inserted to any layer (\ie, $g$) of the pre-trained model. As shown in Fig.~\ref{fig:framework} (b), the adapter is generally a residual block composed of a down-projection with parameters $\mathbf{W}_{down}$, a nonlinear activation function $\sigma(\cdot)$, and an up-projection with parameters $\mathbf{W}_{up}$. The two projections can be convolution~\cite{res_adapter} for CNN or linear~\cite{adapter_tuning} layers  for Transformer architectures, respectively. We formulate the adapter as follows:
\begin{equation}
  \label{eq:seq_adapter}
  \mathbf{h}' = \mathbf{h} + \sigma(\mathbf{h} \ast \mathbf{W}_{down}) \ast \mathbf{W}_{up},
\end{equation}
where the $\ast$ is matrix multiplication or convolution operation, $\sigma$ is the activation function.
Alternatively, the adapter can also be parallel with $g$ like a residual branch~\cite{side_tuning, uniview_pet}:
\begin{equation}
  \label{eq:parallel_adapter}
  \mathbf{h}' = \mathbf{h} + \sigma(\textbf{e} \ast \mathbf{W}_{down}) \ast \mathbf{W}_{up}.
\end{equation}
\smallskip
\noindent\textbf{LoRA}~\cite{lora}
assumes the change of parameters is in a low-rank space when tuning the pre-trained model on a downstream task. For a linear layer with weight $\mathbf{W} \in \mathbb{R}^{d \times d'}$, the weight updates $\Delta \mathbf{W}$ can be decomposed into the multiplication of two small matrices:
\begin{equation}
  \Delta \mathbf{W} = \mathbf{W}_{down} \mathbf{W}_{up},
\end{equation}
where $\mathbf{W}_{down} \in \mathbb{R}^{d \times r}$ and $\mathbf{W}_{up} \in \mathbb{R}^{r \times d'}$. For the convolution layer, the updates can be reshaped into the kernel shape.
Finally, LoRA modifies the forward pass of the adapted layer into the following form:
\begin{equation}
  \label{eq:lora}
  \textbf{h}' = \textbf{h} + \textbf{e} \ast (\mathbf{W}_{down} \mathbf{W}_{up}),
\end{equation}
where $\ast$ is matrix multiplication or convolution operation,
the bias and reshape operation are omitted for conciseness. Since LoRA adapts the weight of $g$, the weight updates can be merged into $g$ to reduce the inference latency.

\smallskip
\noindent\textbf{Prefix}~\cite{prefix_tuning} and \textbf{Prompt}~\cite{prompt_tuning} are learnable tokens prepended to the input of a transformer block or keys and values of the attention module. Given two sets of prefix tokens $\mathbf{P}_k, \mathbf{P}_v \in \mathbb{R}^{l \times d}$ the attention module is modified as:
\begin{equation}
  \begin{aligned}
    \mathbf{h}' = \operatorname{Attn}\left(
    \mathbf{x} \mathbf{W}_q,
    [\mathbf{P}_k, \mathbf{e} \mathbf{W}_k],
    [\mathbf{P}_v, \mathbf{e} \mathbf{W}_v]
    \right),
  \end{aligned}
\end{equation}
where $[\cdot, \cdot]$ is  concatenate, and $\operatorname{Attn}$ is defined as:
$$
  \operatorname{Attn}\left(\mathbf{Q}, \mathbf{K}, \mathbf{V} \right) := \operatorname{softmax}\left(\frac{\mathbf{Q} \mathbf{K}^T}{\sqrt{d}}\right) \mathbf{V},
$$
and the multi-head mechanism is omitted for conciseness.

In this paper, we follow \cite{uniview_pet} to add a learnable scale parameter $s$ to the parallel Adapter (Eq.~\ref{eq:parallel_adapter}) and LoRA (Eq.~\ref{eq:lora}) respectively to obtain:
\begin{equation}
  \label{eq:sp_adapter}
  \mathbf{h}' = \mathbf{h} + s \cdot \sigma(\mathbf{e} \ast \mathbf{W}_{down}) \ast \mathbf{W}_{up},
\end{equation}
%--------
\begin{equation}
  \label{eq:sp_lora}
  \mathbf{h}' = \mathbf{h} + s \cdot \mathbf{e} \ast (\mathbf{W}_{down} \mathbf{W}_{up}).
\end{equation}
The Eqs. (\ref{eq:sp_adapter}) and (\ref{eq:sp_lora}) are general forms of Eqs. (\ref{eq:parallel_adapter}) and (\ref{eq:lora}), and they degenerate to Eq. (\ref{eq:parallel_adapter}) and (\ref{eq:lora}) when $s$ is the constant $1$. We use these PETs in Eqs.~(\ref{eq:sp_adapter}) and~(\ref{eq:sp_lora}) rather than Eqs.~(\ref{eq:seq_adapter}) and~(\ref{eq:lora}) in our experiments (Sec.~\ref{sec:experiment}).

In addition to the three PET modules described above, there are many others, such as AdaptBias~\cite{adaptbias}, Compacter~\cite{compacter}, and AdapterFormer~\cite{adapter_former}, and there will be new and superior PET methods in future as well. All of them can be applied to our CL framework (see Sec.~\ref{sec:framework}), as long as they are compatible with the pre-trained model.

\section{Methodology}

\subsection{Naive Baseline}
We construct a baseline by leveraging pre-trained models and PET techniques, following the naive sequential fine-tuning (Seq-FT) that was usually considered as the lower bound of CIL. Intuitively, our baseline creates a PET module and attaches it to the pre-trained model, then sequentially learns tasks in the same way as Seq-FT but keeping the pre-trained model frozen. We chose the local cross-entropy loss (CE) rather than the global CE loss as the learning objective for Seq-FT and our baseline because local CE empirically performs better than global CE when using a large pre-trained model~\cite{l2p,dual_prompt,esn}. The local CE is the standard CE computed on categories of the current task:
\begin{equation}
  \label{eq:loss_lce}
  \mathcal{L} = \frac{1}{|\mathcal{D}_i|} \sum_{(\mathbf{x}, y) \in \mathcal{D}_i} \mathcal{L}_{ce}(\operatorname{mask}(f(\mathbf{x};\boldsymbol{\theta}, \boldsymbol{\phi})), y),
\end{equation}
where $y$ is the ground truth label of the input $\mathbf{x}$ in the current training set $\mathcal{D}_i$, $\operatorname{mask}(\cdot)$ is a function that filters out the logits of old categories. The Eq.~(\ref{eq:loss_lce}) falls back to global CE when $\operatorname{mask}(\cdot)$ is removed. Although our baseline is very naive, the performance is comparable to the state-of-the-art DualPrompt when using the same Prefix module.

\subsection{Proposed Framework}
\label{sec:framework}
LAE builds upon our network architecture generalizable baseline and additionally introduces three novel designs, yielding a robust framework that can readily reshape any PET methods into a competitive Continual Learning approach. In the following, we will delve into these three key aspects of LAE: learning, accumulation, and ensemble.

\smallskip
\noindent\textbf{Learning with calibrated speed}.
We observed that \textit{PET modules vary in their speed for acquiring new knowledge, leading to disparities in performance}.
In theory, adapting the PET module to a new task too quickly can lead to overfitting and result in worse catastrophic forgetting, whereas slower adaptation can maintain the model's stability but limit its plasticity. We argue that aligning the adaptation speeds of different PET modules is crucial for transforming them into an efficient and robust CL approach in a unified way. To address this, we propose calibrating PET modules to align their adaptation speeds.

Moreover, as the same PET module $\boldsymbol{\theta}_{pet}$ is shared by new and old tasks, the changes made to $\boldsymbol{\theta}_{pet}$ for the new task can cause forgetting for the old tasks. Therefore, slowing down the change in $\boldsymbol{\theta}_{pet}$, \eg, by reducing the learning rate (see supplementary), can help mitigate catastrophic forgetting.
Kumar et al.~\cite{lpft} showcase how the linear probing followed by fine-tuning strategy effectively balances performance across out-of-distribution and in-distribution tasks. We employ a similar technique to calibrate the $\boldsymbol{\theta}_{pet}$'s adaptation speed relative to $\boldsymbol{\phi}_{new}$.
Specifically, at the beginning of the training, we only learn $\boldsymbol{\phi}_{new}$ with $\boldsymbol{\theta}_{pet}$ frozen; then after $\boldsymbol{\phi}_{new}$ has sufficiently learned and the loss has significantly decreased, we jointly learn both $\boldsymbol{\phi}_{new}$ and $\boldsymbol{\theta}_{pet}$.

According to the study by He et al.~\cite{uniview_pet}, the Prefix can be equivalently transformed into a similar form to the Adapter:
\begin{equation}
  \label{eq:prefix_uni_form}
  \mathbf{h}’ \leftarrow(1-\lambda(\mathbf{e}))
  \mathbf{h} + \lambda(\mathbf{e}) \sigma\left(
  \mathbf{e} \mathbf{W}_{1}
  \right) \mathbf{W}_{2},
\end{equation}
in which $\mathbf{W}_{1}$$=$$\mathbf{W}_q \mathbf{P}_k^{\top}$, $\mathbf{W}_{2}$$=$$\mathbf{P}_v$, $\sigma$$=$$\text{softmax}$ and
\begin{equation}
  \label{eq:lambda_x}
  \lambda(\mathbf{e}) = \frac{\sum_i \exp \left(\mathbf{e} \mathbf{W}_q \mathbf{P}_k^{\top}\right)_i}{\sum_i \exp \left(\mathbf{e} \mathbf{W}_q \mathbf{P}_k^{\top}\right)_i + \sum_j \exp \left(\mathbf{e} \mathbf{W}_q \mathbf{W}_k^{\top} \mathbf{C}^{\top}\right)_j}.
\end{equation}
As $\mathbf{P}_k$ usually contains much fewer tokens than input $\mathbf{C}$, $\lambda(\mathbf{e})$ is often a small positive number close to 0, which impacts the gradient of Prefix tokens $\mathbf{P}_v$:
\begin{equation}
  \label{eq:prefix_grad}
  \frac{\partial{\mathcal{L}}}{\partial{\mathbf{P}_v}} =
  (\frac{\partial{\mathbf{h}’}}{\partial{\mathbf{P}_v}})^{\top}
  \frac{\partial{\mathcal{L}}}{\partial{\mathbf{h}’}}
  = \lambda(\mathbf{e}) (\sigma\left(\mathbf{e} \mathbf{W}_{1} \right))^{\top} \frac{\partial{\mathcal{L}}}{\partial{\mathbf{h}’}}.
\end{equation}

Therefore, the gradient of $\mathbf{P}_v$ is significantly smaller than $\mathbf{W}_{up}$ of the corresponding Adapter parameterized by $\mathbf{W}_{down}$$=$$\mathbf{W}_{1}$ and $\mathbf{W}_{up}$$=$$\mathbf{W}_{2}$,
and we can arrive at a similar conclusion regarding $\mathbf{P}_k$ and $\mathbf{W}_{down}$. Then, we can easily observe that Prefix adapts to the new task much slower than Adapter. This is partly why the prompts for different tasks in prior approaches~\cite{l2p, dual_prompt} are prone to be homogeneous (see supplementary). Here we align Prefix with Adapter by compensating its gradient by $\frac{1}{\lambda(\mathbf{e})}$ and adding two learnable scaling parameters $s^k$ and $s^v$, calibrating the Prefix described by Eq.~(\ref{eq:prefix_uni_form}) into the following form:
\begin{equation}
  \label{eq:prefix_uni_form_improved}
  \mathbf{h}’ \leftarrow(1-\lambda(\mathbf{e}))
  \mathbf{h} + \sigma\left(
  s^k \cdot \mathbf{e} \mathbf{W}_{1}
  \right) (s^v \cdot \mathbf{W}_{2}).
\end{equation}
The adaptation speed of the calibrated Prefix is nearly equivalent to the Adapter depicted in Eq.~(\ref{eq:sp_adapter}) and elevates its performance to be on par with the Adapter. For other PET modules, we can also analyze them specifically and then calibrate their adaptation speeds to align with Adapter.

By aligning the adaptation speed of PET modules and calibrating their adaptation speed relative to the classifiers, our framework achieves a better and more consistent stability-plasticity balance with various PET modules.

\smallskip
\noindent\textbf{Accumulation of multi-task knowledge}.
The PET module $\boldsymbol{\theta}_{pet}$ is designed to continuously adapt to new tasks, making the model more proficient in dealing with novel tasks. However, this adaptation process can result in the model gradually forgetting how to handle older tasks. To address this issue, we propose to create an additional expert for older tasks to complement the expert for newer tasks, drawing inspiration from the Complementary Learning System~\cite{cls_theory, cls_theory_updated} of the human brain, which involves the hippocampus rapidly learning new knowledge and the neocortex integrating learned knowledge in an offline manner over time. We achieve this by duplicating the online PET module $\boldsymbol{\theta}_{pet}^{on}$ (\emph{i.e.}, the $\boldsymbol{\theta}_{pet}$ in the baseline) attached to the model as the offline PET module $\boldsymbol{\theta}_{pet}^{off}$ after the model has learned the first task. The $\boldsymbol{\theta}_{pet}^{off}$ slowly accumulates the learned knowledge when the model learns a new task by an accumulation function, and we empirically find the simple Exponential Moving Average (EMA) algorithm works well for our LAE:
\begin{equation}
  \label{eq:ema}
  \boldsymbol{\theta}_{pet}^{off} \leftarrow \alpha \cdot \boldsymbol{\theta}_{pet}^{off} + (1 - \alpha) \cdot \boldsymbol{\theta}_{pet}^{on},
\end{equation}
where $\alpha \in (0, 1)$ is a large (\emph{i.e.}, close to 1) weight decay.

This way, the neocortex-like offline PET module gradually integrates the learned knowledge in a slow offline manner, while the hippocampus-like online PET module continues to rapidly learn new knowledge. Then, we can obtain two experts for newer tasks and older tasks with $\boldsymbol{\theta}_{pet}^{on}$ and $\boldsymbol{\theta}_{pet}^{off}$, respectively. However, the task a sample belongs to is unknown during inference, we need to devise a method to effectively utilize both experts for inference.

\smallskip
\noindent\textbf{Ensemble of two expert models}.
Two expert models constructed with $\boldsymbol{\theta}_{pet}^{on}$ and $\boldsymbol{\theta}_{pet}^{off}$ are respectively proficient at handling newer and older tasks. Instead of inference only using the online or offline expert model, we integrate their outputs to obtain the prediction for an inference sample.

A classifier can be viewed as an energy model when we define the unnormalized negative log probability as the energy function~\cite{ebm_tutorial}. The optimization goal of the energy model is to minimize the energy of the model on the data distribution of its learning task. Previous research~\cite{eb_ood} has shown that the energy of an energy model trained on one data domain is generally very high on other data domains. Therefore, the Eq.~(\ref{eq:loss_lce}) actually continuously minimizes the energy of the $\boldsymbol{\phi}_{new}$  on the new task. Even if old data is not used during training, the energy of the old data on $\boldsymbol{\phi}_{new}$ will be very high, as demonstrated by the recent work ESN~\cite{esn}.

Likewise, as the $\boldsymbol{\theta}_{pet}^{on}$ and $\boldsymbol{\theta}_{pet}^{off}$ respectively contain relatively more novel and historical knowledge, theoretically, the energy produced by $\boldsymbol{\theta}_{pet}^{on}$ for the sample of the newer tasks should be smaller than that produced by $\boldsymbol{\theta}_{pet}^{off}$, and the vice versa for the sample of older tasks. Therefore, choosing the prediction result with the lowest energy as the final prediction of an inference sample seems like a simple but effective solution. However, in practice, we find that normalizing the energy produced by $\boldsymbol{\theta}_{pet}^{on}$ and $\boldsymbol{\theta}_{pet}^{off}$ before ensemble yields more robust results. Therefore, we adopt the following ensemble algorithm instead:
\begin{equation}
  \label{eq:ensemble_method}
  f_{ens}(\mathbf{o}^{on}, \mathbf{o}^{off}) := \operatorname{max}\left(
  \operatorname{\sigma}\left(\mathbf{o}^{on}\right),
  \operatorname{\sigma}\left(\mathbf{o}^{off}\right)
  \right),
\end{equation}
where $\operatorname{\sigma}$ is the softmax function, $\mathbf{o}^{on}$ and $\mathbf{o}^{off}$ are outputs of the online and offline expert models (\emph{i.e.}, $f(\cdot;\boldsymbol{\theta}_{pet}^{on}, \boldsymbol{\phi})$ and $f(\cdot;\boldsymbol{\theta}_{pet}^{off}, \boldsymbol{\phi})$) for an inference sample, respectively.

\smallskip
As illustrated in Fig.~\ref{fig:framework}, in our LAE framework, the model learns a new task with $\boldsymbol{\theta}_{pet}^{on}$ and accumulates the learned knowledge to $\boldsymbol{\theta}_{pet}^{off}$, the two experts favored by newer and older tasks are ensembled to get the final prediction for an inference sample. Our LAE can be applied to the pre-trained model in any network architecture as long as the PET modules are compatible with the model.

\section{Experiment}
\label{sec:experiment}

\subsection{Datasets and Evaluation Protocols}
Our experiments use models pre-trained on the ImageNet21k~\cite{imagenet} dataset without specified, and we follow prior works to train and evaluate the model on CIFAR100~\cite{cifar} and ImageNet-R~\cite{imagenet-r} benchmarks.

\noindent\textbf{CIFAR100} is an extensively used dataset in prior continual learning (CL) works, containing 100 classes, each class with 500 training and 100 test images of size 32$\times$32$\times$3.

\noindent\textbf{ImageNet-R} is first introduced to CL by Wang et al.~\cite{dual_prompt}, including 200 subcategories of ImageNet~\cite{imagenet}, but its samples are in different styles, such as cartoon, graffiti, and origami. There are also some hard examples from ImageNet that standard models, \emph{e.g.}, ResNet~\cite{resnet}, fail to classify. The original dataset is split into the train set with 24000 samples and the test set with 6000 samples, and the number of training and testing samples varies between classes.

We follow prior works to split the dataset into 10 tasks, and all tasks have the same number of classes, \emph{i.e.}, 10 for CIFAR100 and 20 for ImageNet-R. We evaluate the model by the widely used incremental metrics: last incremental accuracy $A_{N}$ and average incremental accuracy $\bar{A}_{N} = \frac{1}{N} \sum_{i=1}^N A_i$, where $N$ is the total number of tasks (\emph{i.e.}, 10), and $A_i$ is formally defined as:
\begin{equation}
  A_i = \frac{1}{\lvert \mathcal{D}_{1:i}^{test} \rvert}
  \sum_{(\mathbf{x}, y) \in \mathcal{D}_{1:i}^{test}}
  \mathds{1} \left(\hat{y} = y\right),
\end{equation}
where $\mathds{1}(\cdot)$ is the indicator function that maps the boolean value to $\{0, 1\}$, $\mathcal{D}_{1:i}^{test}$ is the test set of all seen tasks so far, $\hat{y}$ and $y$ are predicted and ground truth labels of input $\mathbf{x}$. \textit{We ran all experiments 3 times with different class orders and report the mean and standard deviation of these 3 runs}.

\begin{table}
  \renewcommand\arraystretch{0.85}
  \caption{Benchmark Results on CIFAR100. The PET modules are inserted into the first 5 transformer blocks of the standard ViT-B/16 pre-trained on the ImageNet21k dataset. The ``5, 10, 20'' indicate the size of PET modules.}
  \vskip -0.25in
  \label{table:cifar100}
  \begin{center}
    \resizebox{\columnwidth}{!}{
      \begin{tabular}{lcccr}
        \toprule
        Approach                      & PET Module & $A_{10}$ (↑)            & $\bar{A}_{10}$ (↑)      \\
        \midrule
        Joint-FT                      & -          & 92.00$\pm$0.18          & -                       \\
        Seq-FT                        & -          & 77.61$\pm$0.37          & 85.82$\pm$0.86          \\
        \midrule
        \multirow{6}{*}{Baseline}     & Adapter5   & 82.30$\pm$1.20          & 88.18$\pm$0.31          \\
                                      & Adapter10  & 81.76$\pm$1.21          & 87.84$\pm$0.49          \\
                                      & LoRA5      & 83.24$\pm$1.54          & 88.48$\pm$0.31          \\
                                      & LoRA10     & 82.55$\pm$1.61          & 88.35$\pm$0.48          \\
                                      & Prefix10   & 84.49$\pm$0.30          & 89.34$\pm$0.59          \\
                                      & Prefix20   & 84.44$\pm$0.75          & 89.46$\pm$0.42          \\
        \midrule
        \midrule
        L2P~\cite{l2p}                & Prompt     & 82.57$\pm$0.42          & 86.95$\pm$0.68          \\
        DualPrompt~\cite{dual_prompt} & Prefix20   & 84.27$\pm$0.41          & 88.92$\pm$0.78          \\
        ESN~\cite{esn}                & Prompt     & 84.18$\pm$0.08          & 88.49$\pm$0.64          \\
        \midrule
        \multirow{6}{*}{LAE (Ours)}   & Adapter5   & \textbf{85.59}$\pm$0.46 & \textbf{89.96}$\pm$0.44 \\
                                      & Adapter10  & 85.33$\pm$0.20          & 89.77$\pm$0.50          \\
                                      & LoRA5      & 85.56$\pm$0.16          & 89.63$\pm$0.41          \\
                                      & LoRA10     & 85.37$\pm$0.39          & 89.87$\pm$0.50          \\
                                      & Prefix10   & 85.17$\pm$0.14          & 89.73$\pm$0.43          \\
                                      & Prefix20   & 85.25$\pm$0.66          & 89.71$\pm$0.42          \\
        \bottomrule
      \end{tabular}
    }
  \end{center}
  \vskip -0.3in
\end{table}

\subsection{Implementation and Training Details}
To make fair comparisons, we consider state-of-the-art approaches~\cite{l2p, dual_prompt} based on pre-trained models like our LAE and using the PyTorch code released by Jaeho Lee\footnote{https://github.com/JH-LEE-KR} to conduct experiments. The joint fine-tuning (Joint-FT) and the naive sequential fine-tuning (Seq-FT) usually recognized as the upper and lower bounds of CIL are implemented in our codebase, referring to the code of Jaeho Lee. We also compare with recent work ESN~\cite{esn}, using its official PyTorch code. We chose three types of representative PET modules and two sizes per type for our baseline and LAE framework, where \textit{the size denotes the down-projection dimension of the Adapter, the rank of LoRA, or the length of the Prefix} described in Sec.~\ref{sec:preliminaries}. We assume that only a single PET module is attached to the pre-trained model in the previous discussion for convenience, in practice, multiple PET modules are inserted into the Attention blocks of Transformers or the convolution blocks of ConvNets in the shallow layers, following DualPrompt~\cite{dual_prompt}.

The training strategy of our baseline and LAE framework is the same as DualPrompt, \emph{i.e.}, training the model with Adam optimizer for 5 and 50 epochs, and constant learning rate 0.03 and 0.005 based on batch size 256, for CIFAR100 and ImageNet-R, respectively. The EMA algorithm's weight decay $\alpha$ defined in Eq.~(\ref{eq:ema}) is empirically set to 0.9999 in all experiments. The freezing epochs of PET modules are set to 3 and 30 for CIFAR100 and ImageNet-R, respectively.
The data augmentation is consistent with that used in model pre-training. We train Joint-FT and Seq-FT with the recommended fine-tuning strategy of ViT~\cite{vit}, but the number of training epochs is the same as ours. More details can be found in the supplementary materials.

\begin{table}
  \renewcommand\arraystretch{0.85}
  \caption{Benchmark Results on ImageNet-R. The PET modules are inserted into the first 5 transformer blocks of the standard ViT-B/16 pre-trained on the ImageNet21k dataset. The ``5, 10, 20'' indicate the size of PET modules.}
  \vskip -0.25in
  \label{table:imagenet-r}
  \begin{center}
    \resizebox{\columnwidth}{!}{
      \begin{tabular}{lcccr}
        \toprule
        Approach                      & PET Module & $A_{10}$ (↑)            & $\bar{A}_{10}$ (↑)      \\
        \midrule
        Joint-FT                      & -          & 79.69$\pm$0.16          & -                       \\
        Seq-FT                        & -          & 40.42$\pm$3.28          & 59.39$\pm$2.36          \\
        \midrule
        \multirow{6}{*}{Baseline}     & Adapter5   & 61.63$\pm$2.51          & 71.58$\pm$2.33          \\
                                      & Adapter10  & 57.08$\pm$3.67          & 68.58$\pm$2.89          \\
                                      & LoRA5      & 60.79$\pm$2.63          & 70.50$\pm$2.23          \\
                                      & LoRA10     & 57.62$\pm$3.74          & 68.21$\pm$2.85          \\
                                      & Prefix10   & 68.94$\pm$1.25          & 75.31$\pm$1.51          \\
                                      & Prefix20   & 68.99$\pm$0.98          & 75.38$\pm$1.43          \\
        \midrule
        \midrule
        L2P~\cite{l2p}                & Prompt     & 63.91$\pm$1.60          & 69.27$\pm$2.25          \\
        DualPrompt~\cite{dual_prompt} & Prefix20   & 68.99$\pm$0.08          & 74.21$\pm$1.13          \\
        ESN~\cite{esn}                & Prompt     & 62.61$\pm$0.96          & 68.58$\pm$1.64          \\
        \midrule
        \multirow{6}{*}{LAE (Ours)}   & Adapter5   & \textbf{72.66}$\pm$0.63 & 78.91$\pm$0.89          \\
                                      & Adapter10  & 72.45$\pm$0.81          & \textbf{79.07}$\pm$0.88 \\
                                      & LoRA5      & 72.00$\pm$0.75          & 78.33$\pm$0.96          \\
                                      & LoRA10     & 71.83$\pm$0.57          & 78.24$\pm$0.90          \\
                                      & Prefix10   & 71.85$\pm$0.66          & 77.44 $\pm$1.12         \\
                                      & Prefix20   & 72.05$\pm$0.66          & 77.55$\pm$1.00          \\
        \bottomrule
      \end{tabular}
    }
  \end{center}
  \vskip -0.3in
\end{table}

\subsection{Benchmark Results}
\noindent\textbf{CIFAR100} benchmark results are present in Tab.~\ref{table:cifar100}. All approaches use the same ViT-B/16~\cite{vit} model pre-trained on the ImageNet21k~\cite{imagenet} dataset. The numerical suffix of the PET module denotes its size (\emph{i.e.}, down-projection dimension or length). L2P and DualPrompt are state-of-the-art approaches that adopt a pool to store Prompt or Prefix. However, the accuracy of their prompt selection gradually declines with the increase in the number of learning tasks and the prompts for different tasks appear homogeneous (see supplementary). Therefore, our baseline is very naive but achieves comparable performance to L2P and DualPrompt, and our LAE framework with all 6 PET modules consistently surpasses DualPrompt and ESN by about 1.5\% in last incremental accuracy $A_{10}$.
Although PET modules have different performances in the baseline, they achieve better and same level performance in our LAE, mainly due to the calibration of adaptation speed. In particular, DualPrompt has 3-10x more learnable parameters than our LAE.

\begin{figure}[t]
  \begin{center}
    \begin{subfigure}[b]{0.49\columnwidth}
      \centering
      \includegraphics[width=\textwidth]{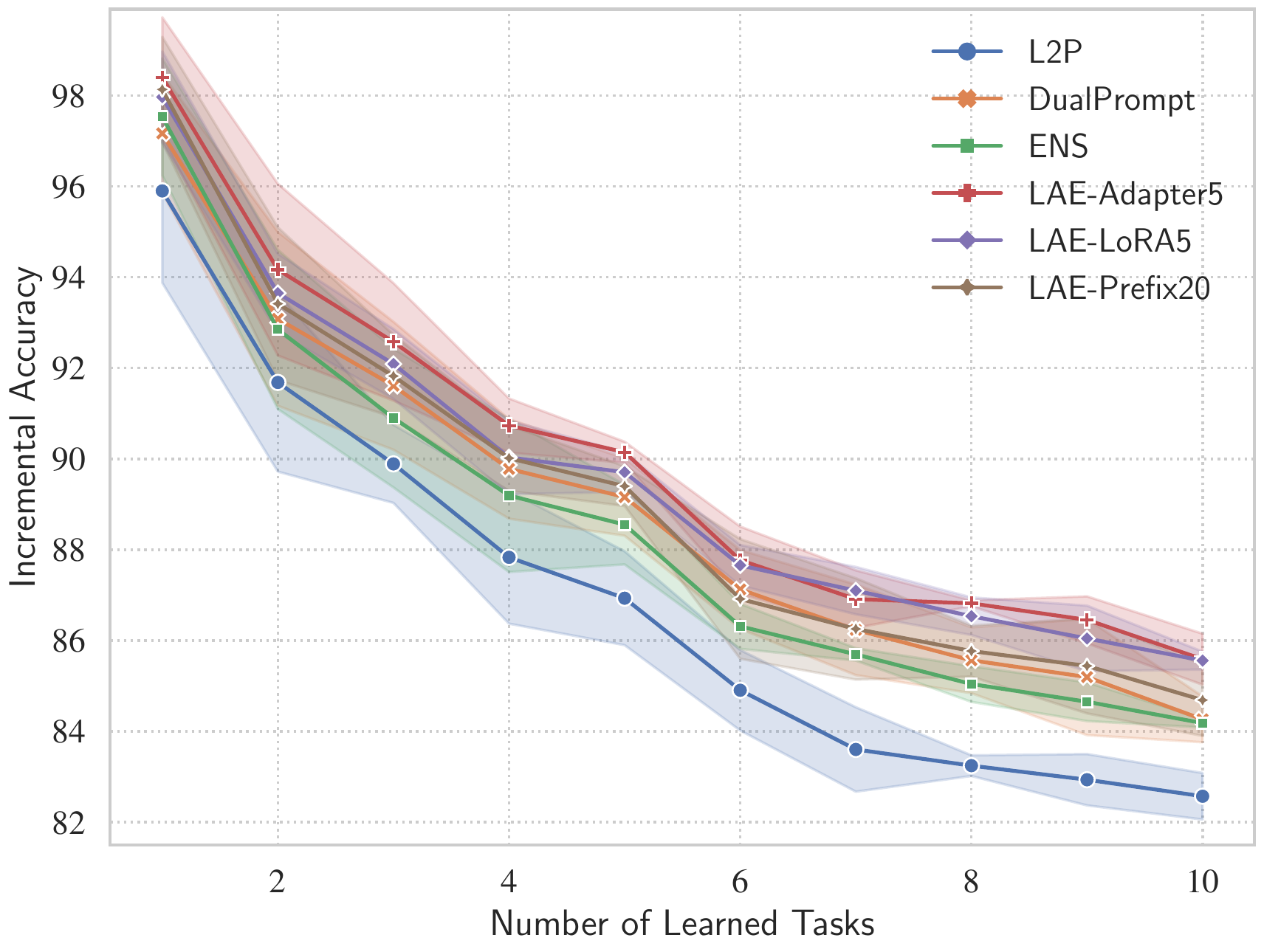}
      \caption{CIFAR100}
      \label{fig:cifar_benchmark}
    \end{subfigure}
    \hfill
    \begin{subfigure}[b]{0.49\columnwidth}
      \centering
      \includegraphics[width=\textwidth]{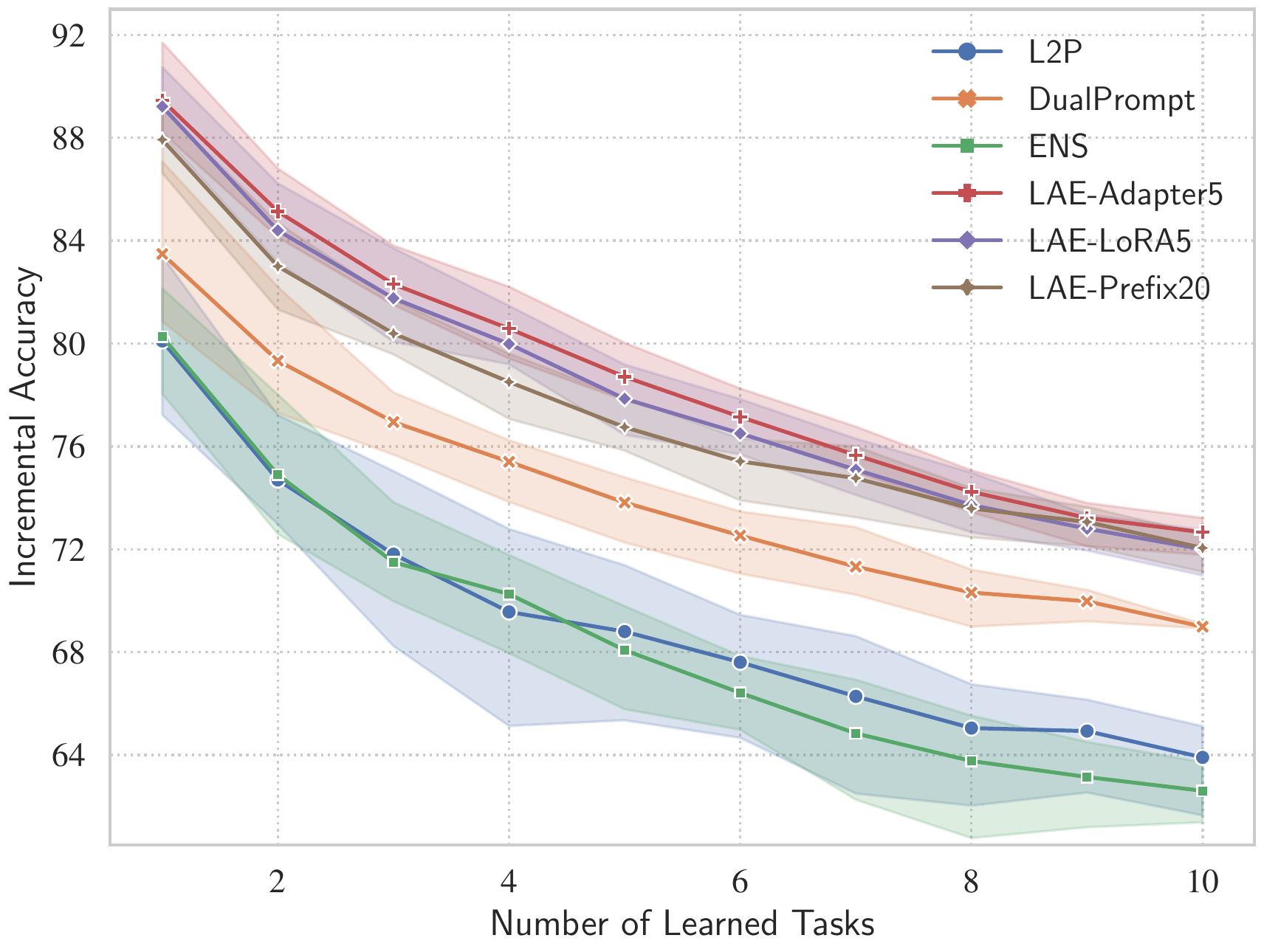}
      \caption{ImageNet-R}
      \label{fig:imgr_benchmark}
    \end{subfigure}
    \hfill
    \vskip -0.1in
    \caption{Task-by-Task Incremental Accuracy on two benchmarks. The lines show the task-by-task evaluation results of L2P~\cite{l2p}, DualPrompt~\cite{dual_prompt}, ENS~\cite{esn}, and our LAE framework with different PET modules.}
    \label{fig:benchmarks}
  \end{center}
  \vskip -0.35in
\end{figure}

\noindent\textbf{ImageNet-R} benchmark is more difficult than CIFAR100, but it can better demonstrate the advantages of our LAE framework. From the results shown in Tab.~\ref{table:imagenet-r}, our baseline can only achieve comparable performance to DualPrompt when using Prefix. This is because the Adapter and LoRA adapt to a new task faster than Prefix, which is enlarged in the ImageNet-R dataset but successfully addressed by the adaption speed calibration of our LAE framework. Thus, we can see that our LAE achieves more than 3.5\% performance improvement over DualPrompt in terms of the last incremental accuracy $A_{10}$, which is also scaled up compared to the easier CIFAR100 dataset. We can also observe that the size of the PET modules has little impact on performance in our LAE framework, and our LAE is more robust to the class order, while our baseline has a relatively large variance between different class orders.

The real-world CL is an endless procedure, and the performance of each learning phase is equally important to the AI system. So, we also plot the task-by-task incremental accuracy in Fig.~\ref{fig:cifar_benchmark} and \ref{fig:imgr_benchmark}. We can observe that our LAE with all three types of PET modules performs better than L2P and DualPrompt at almost all learning phases. \textit{Our LAE outperforms others by a wider margin in the 20-task experiments presented in the supplementary material, highlighting its ability to handle long-term CL scenarios}. In the supplementary, we compare our LAE with the contemporary CODA-Prompt~\cite{coda_prompt} approach on ImageNet-R and DomainNet~\cite{domainnet} datasets. The parameters and computation comparison can also be found in the supplementary.

\subsection{Ablation Study}
\begin{figure}[t]
  \begin{center}
    \begin{subfigure}[b]{0.49\columnwidth}
      \centering
      \includegraphics[width=\textwidth]{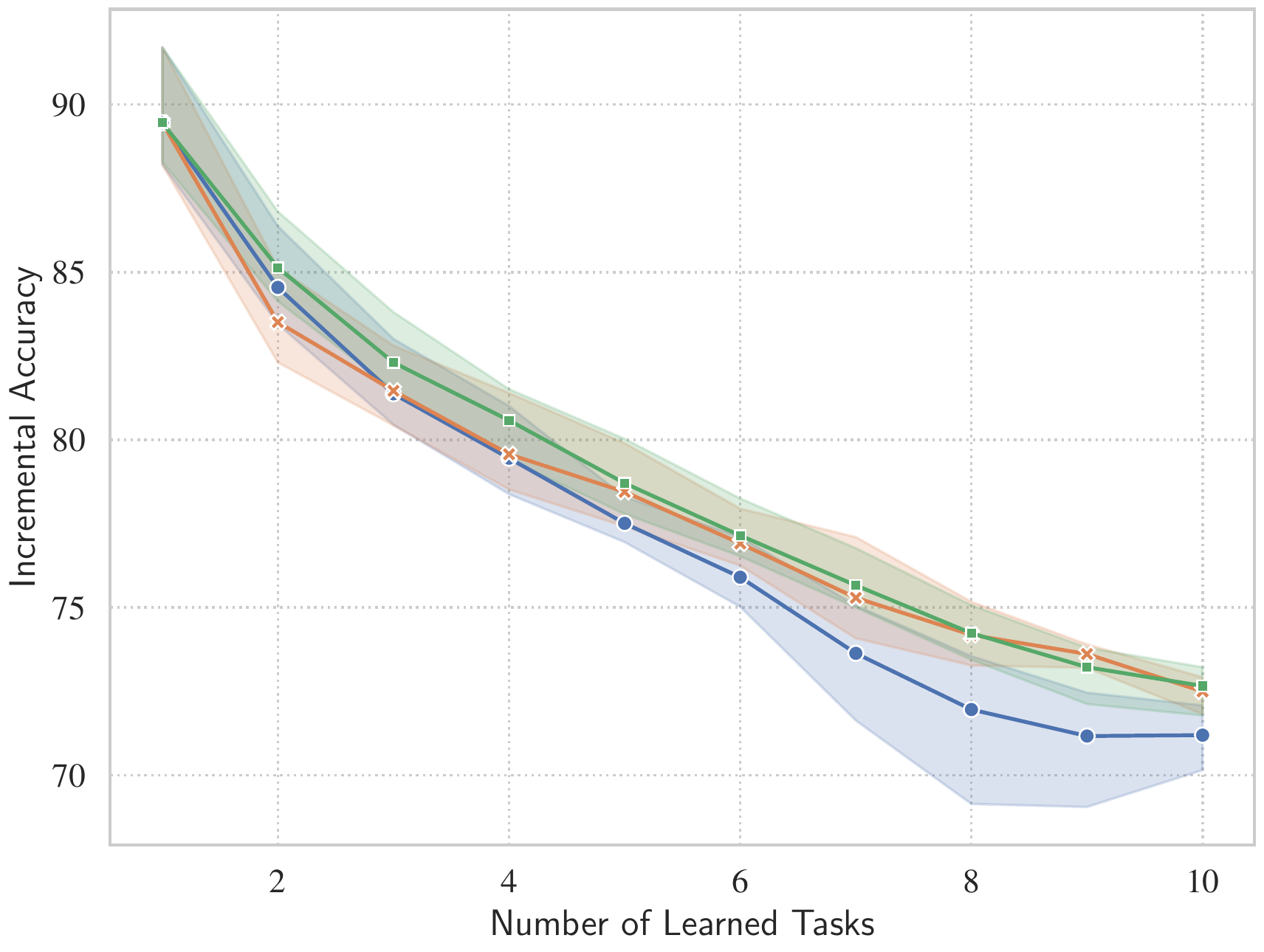}
      \caption{Adapter5}
    \end{subfigure}
    \hfill
    \begin{subfigure}[b]{0.49\columnwidth}
      \centering
      \includegraphics[width=\textwidth]{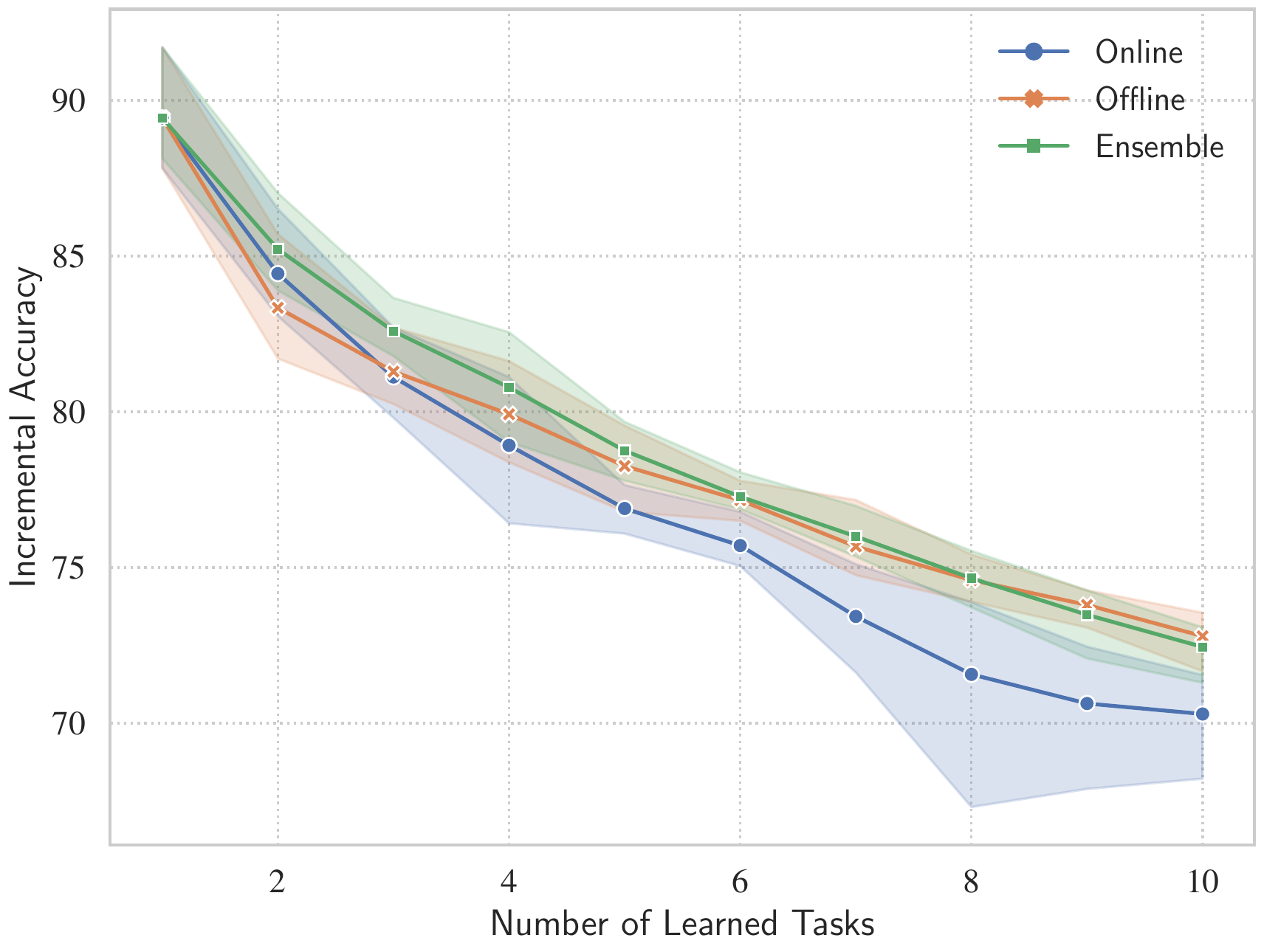}
      \caption{Adapter10}
    \end{subfigure}
    \hfill
    \vskip -0.1in
    \caption{Ablation on inference with Online Adapter, Offline Adapter, and our Experts Ensemble. The three strategies correspond to rows 4, 6, and 7 in Table~\ref{tab:ablation}.}
    \label{fig:ens_ada}
  \end{center}
  \vskip -0.4in
\end{figure}

Our LAE consists of three main novel designs, \emph{i.e.}, learning, accumulation, and ensemble, so we ablate on them and report the results in Tab.~\ref{tab:ablation}. The first and last rows are our baseline and LAE framework, respectively. The performance drops the most when removing our learning with calibrated speed, demonstrating that it contributes the most to our LAE. Accumulation and Ensemble are also important to our LAE, without them the last incremental accuracy decreases by 2.15\%. The sixth row indicates our LAE inference with the Offline PET module only, whose $A_{10}$ is even better than the inference by our expert ensemble. As illustrated in Fig.~\ref{fig:ens_ada}, inference by expert ensemble performs better than inference with the Online or Offline PET module alone in the earlier learning phases. However, as the number of learned tasks increases, the advantage of expert ensemble over the Offline PET module gradually decreases, partly due to the performance of the old tasks dominating the overall performance. Nonetheless, inference with expert ensemble yields more robust performance in most cases.

We also conducted ablation experiments on the calibration made to Prefix. We can see from Tab.~\ref{tab:ablation_prefix} that both gradient compensation and learnable scaling parameters individually lead to significant improvements in performance.
Moreover, when used together, the performance gain is approximately equal to the sum of the gains achieved by using each of them separately, indicating that their contributions to the performance are independent of each other.

\begin{table}
  \renewcommand\arraystretch{0.85}
  \caption{Ablation study on three key designs of our LAE framework: Learning (Lea.), Accumulation (Acc.), and Ensemble (Ens.). The experiments are conducted on the ImageNet-R dataset with Adapter10.}
  \vskip -0.2in
  \label{tab:ablation}
  \begin{center}
    \resizebox{\columnwidth}{!}{
      \begin{tabular}{lcccr}
        \toprule
        Lea.      & Ens.      & Acc.      & $A_{10}$ (↑)            & $\bar{A}_{10}$ (↑)      \\
        \midrule
        \ding{55} & \ding{55} & \ding{55} & 57.08$\pm$3.67          & 68.58$\pm$2.89          \\
        \ding{55} & \ding{51} & \ding{55} & 63.36$\pm$1.46          & 73.78$\pm$1.54          \\
        \ding{55} & \ding{51} & \ding{51} & 65.22$\pm$2.56          & 75.71$\pm$1.18          \\
        \ding{51} & \ding{55} & \ding{55} & 70.30$\pm$1.48          & 77.25$\pm$1.39          \\
        \ding{51} & \ding{51} & \ding{55} & 71.37$\pm$1.24          & 78.27$\pm$1.30          \\
        \ding{51} & \ding{55} & \ding{51} & \textbf{72.80}$\pm$0.81 & 78.63$\pm$0.95          \\
        \ding{51} & \ding{51} & \ding{51} & 72.45$\pm$0.81          & \textbf{79.07}$\pm$0.88 \\
        \bottomrule
      \end{tabular}
    }
  \end{center}
  \vskip -0.2in
\end{table}

\subsection{Attach Position of PET Modules}
\begin{figure}[t]
  \begin{center}
    \begin{subfigure}[b]{0.49\columnwidth}
      \centering
      \includegraphics[width=\textwidth]{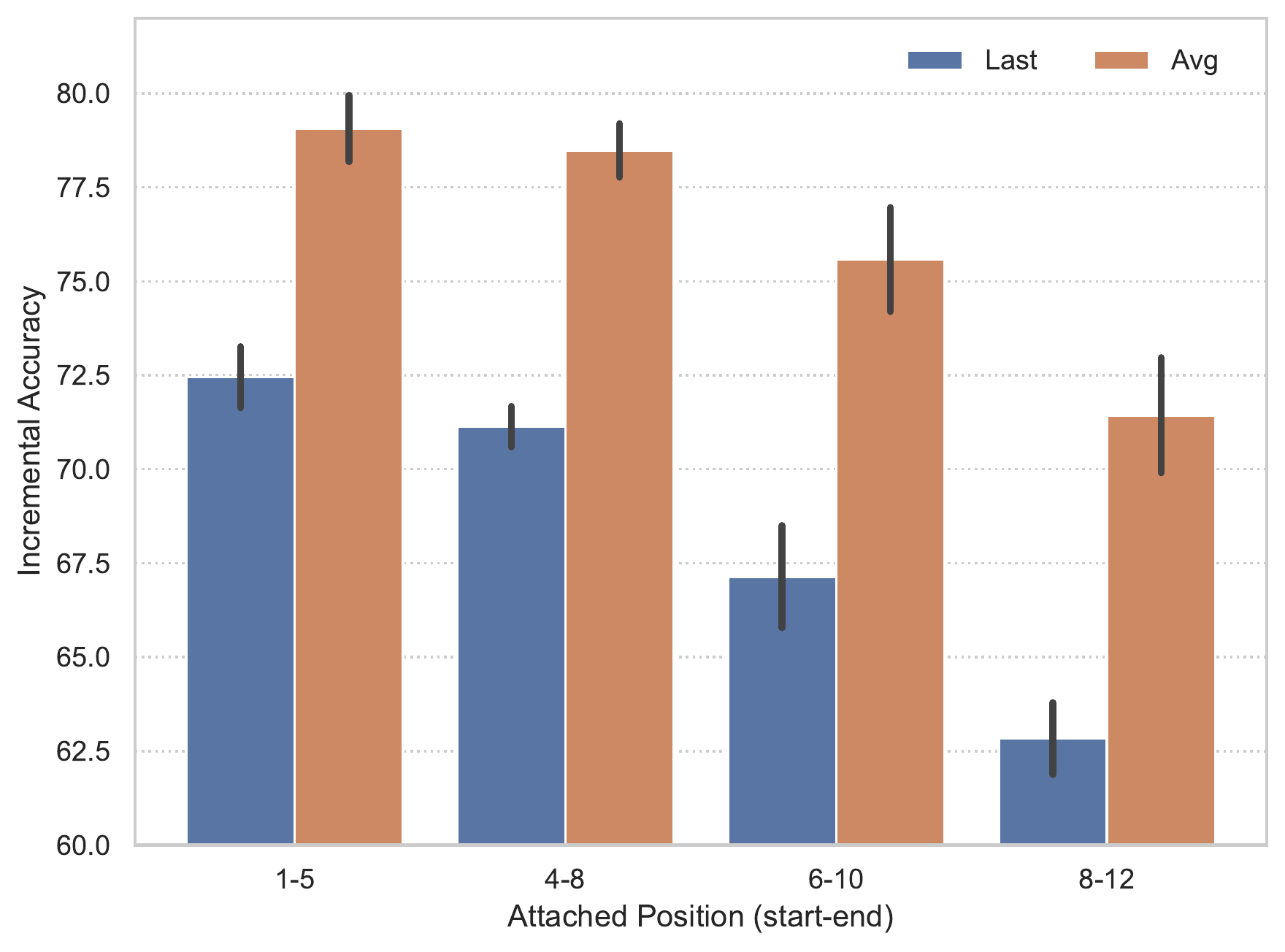}
    \end{subfigure}
    \hfill
    \begin{subfigure}[b]{0.49\columnwidth}
      \centering
      \includegraphics[width=\textwidth]{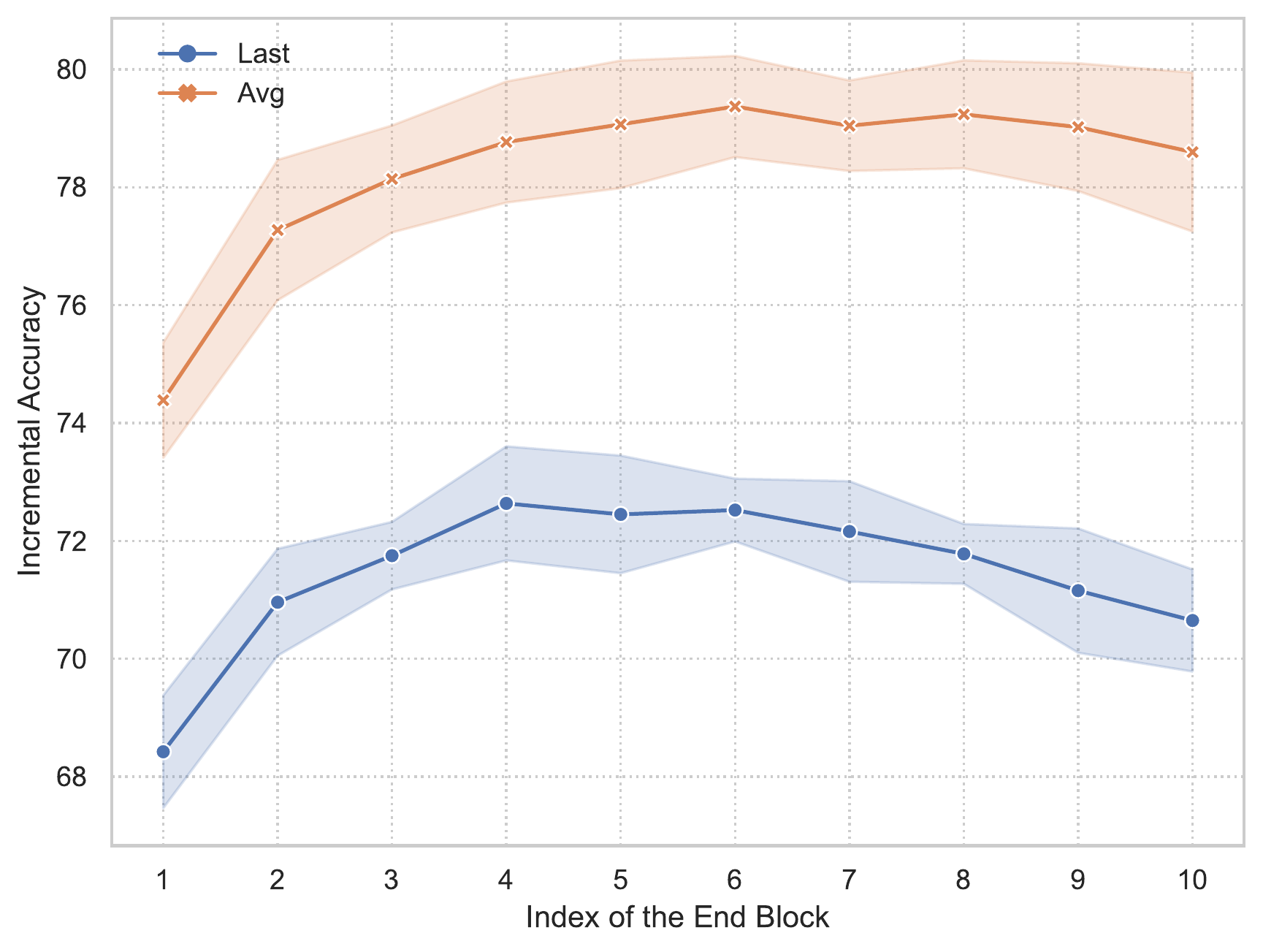}
    \end{subfigure}
    \hfill
    \caption{Ablation on the attached position of PET module (left) and the number of inserted blocks (right). There are 12 transformer blocks in the pre-trained model, ``1-5'' indicates attaching PET modules to the first 5 transformer blocks. The PET modules are inserted into the transformer starting from the first block in the right figure.}
    \label{fig:ablation_position_and_number}
  \end{center}
  \vskip -0.3in
\end{figure}

Our LAE inserts PET modules directly into the first 5 Transformer blocks, following the DualPrompt. As shown in Fig.~\ref{fig:ablation_position_and_number} (left), inserting the PET modules in the shallowest position produces better results than inserting them in deeper positions, which is consistent with the observation in DualPrompt. Additionally, Fig.~\ref{fig:ablation_position_and_number} (right) shows that inserting PET modules in the first 6 Transformer blocks achieves the best performance while inserting them in the first 5 Transformer blocks (\emph{i.e.}, the default setting of our LAE) also results in nearly the same performance.

\begin{table}
  \renewcommand\arraystretch{0.85}
  \caption{Ablation study on the calibration of Prefix: gradient compensation and learnable scaling parameters. The experiments are conducted on ImageNet-R with Prefix20.}
  \vskip -0.2in
  \label{tab:ablation_prefix}
  \begin{center}
    \resizebox{\columnwidth}{!}{
      \begin{tabular}{lcccr}
        \toprule
        Compensation & Scale     & $A_{10}$ (↑)   & $\bar{A}_{10}$ (↑) \\
        \midrule
        \ding{55}    & \ding{55} & 69.23$\pm$0.47 & 75.11$\pm$1.30     \\
        \ding{51}    & \ding{55} & 70.53$\pm$0.55 & 76.16$\pm$1.16     \\
        \ding{55}    & \ding{51} & 70.38$\pm$0.52 & 76.46$\pm$1.25     \\
        \ding{51}    & \ding{51} & 72.05$\pm$0.66 & 77.55$\pm$1.00     \\
        \bottomrule
      \end{tabular}
    }
  \end{center}
  \vskip -0.2in
\end{table}

\subsection{Results on Transformer variant and ConvNet}
Prefix and Prompt are not flexible enough to be applied to ConvNets and Transformer variants, while our LAE is model architecture generalizable due to the ability to leverage various PET modules. We choose the Swin Transformer~\cite{swin} and ConvNeXt~\cite{convnext} to validate our LAE.

\smallskip
\noindent\textbf{Swin Transformer} is a representative window-based Vision Transformer, but L2P and DualPrompt cannot be directly applied to it because the inserted tokens may disrupt the proper division of the windows. Therefore, we only compare our LAE with our baseline when using Swin Transformer, and report the results in Tab.~\ref{table:swin}. We can see that our LAE achieves better performance with Swin-B than with ViT-B/16, which is mainly due to the superior performance of Swin-B. Moreover, our LAE significantly improves the performance of our baseline on all two datasets.

\smallskip
\noindent\textbf{ConvNeXt} is a modern ConvNet for the 2020s that outperforms Swin Transformer by incorporating several novel designs into the standard ResNet~\cite{resnet}. Similarly, we compare our LAE with the baseline in Tab.~\ref{table:convnext}. The Adapter's down and up projections are implemented using $1$$\times$$1$ convolution layers. Both our baseline and LAE achieve significantly better performance using ConvNeXt-B compared to using ViT-B/16 and Swin-B, highlighting LAE's strengths beyond being limited to Transformers. Our LAE consistently improves the performance of our baseline on both datasets.

\begin{table}
  \renewcommand\arraystretch{0.85}
  \caption{The comparison between our LAE framework with our baseline on two datasets using the Swin-B model pre-trained on the ImageNet22k dataset. Adapters are inserted in the shallower 10 of 24 transformer blocks.}
  \vskip -0.2in
  \label{table:swin}
  \begin{center}
    \resizebox{\columnwidth}{!}{
      \begin{tabular}{lcccr}
        \toprule
        Approach   & Dataset                    & $A_{10}$ (↑)            & $\bar{A}_{10}$ (↑)      \\
        \midrule
        Baseline   & \multirow{2}{*}{CIFAR100}  & 85.85$\pm$0.69          & 90.36$\pm$0.67          \\
        LAE (Ours) &                            & \textbf{86.52}$\pm$0.38 & \textbf{90.58}$\pm$0.61 \\
        \midrule
        Baseline   & \multirow{2}{*}{ImageNetR} & 71.81$\pm$1.09          & 78.91$\pm$1.41          \\
        LAE (Ours) &                            & \textbf{73.38}$\pm$0.70 & \textbf{80.01}$\pm$1.51 \\
        \bottomrule
      \end{tabular}
    }
  \end{center}
  \vskip -0.2in
\end{table}

\begin{table}
  \renewcommand\arraystretch{0.85}
  \caption{The comparison between our LAE framework with our baseline on two datasets using the ConvNeXt-B model pre-trained on the ImageNet22k dataset. Adapters are inserted in the shallower 15 of 36 convolution blocks.
  }
  \vskip -0.2in
  \label{table:convnext}
  \begin{center}
    \resizebox{\columnwidth}{!}{
      \begin{tabular}{lcccr}
        \toprule
        Approach   & Dataset                    & $A_{10}$ (↑)            & $\bar{A}_{10}$ (↑)      \\
        \midrule
        Baseline   & \multirow{2}{*}{CIFAR100}  & 86.40$\pm$0.07          & 91.00$\pm$0.31          \\
        LAE (Ours) &                            & \textbf{87.01}$\pm$0.28 & \textbf{91.18}$\pm$0.34 \\
        \midrule
        Baseline   & \multirow{2}{*}{ImageNetR} & 76.35$\pm$1.36          & 82.60$\pm$1.79          \\
        LAE (Ours) &                            & \textbf{78.38}$\pm$0.80 & \textbf{83.95}$\pm$1.16 \\
        \bottomrule
      \end{tabular}
    }
  \end{center}
  \vskip -0.25in
\end{table}

\section{Conclusion}
This paper thoroughly studied the novel Continual Learning (CL) paradigm that starts with a pre-trained model and continuously adapts the model to arriving tasks utilizing general Parameter-Efficient Tuning (PET) methods. We constructed a naive baseline that achieved performance comparable to the prior state-of-the-art approaches. We proposed the Learning-Accumulation-Ensemble (LAE) framework by introducing three novel designs to the baseline. Our LAE can convert any PET method into an efficient CL approach without accessing any old data. We conducted extensive experiments to validate the effectiveness of our LAE, and the results demonstrated that our LAE significantly outperforms the previous state-of-the-art approaches.

\noindent\textbf{Limitations}. There are still some limitations that need to be improved in the future, such as how to accumulate knowledge more efficiently and better ensemble expert models. Moreover, due to the lack of large-scale datasets that do not overlap with the pre-training dataset, our LAE has not been verified in CL scenarios with a larger number of tasks.

\noindent Overall, this paper provides a new solution for Memory-Free CL and offers some theoretical and experimental references for future research on this novel CL paradigm.

{\small
  \bibliographystyle{ieee_fullname}
  \bibliography{egbib}

\begin{thebibliography}{10}\itemsep=-1pt

\bibitem{mas}
Rahaf Aljundi, Francesca Babiloni, Mohamed Elhoseiny, Marcus Rohrbach, and
  Tinne Tuytelaars.
\newblock Memory aware synapses: Learning what (not) to forget.
\newblock In {\em Proceedings of the European Conference on Computer Vision
  (ECCV)}, 2018.

\bibitem{alssum2023smile}
Lama Alssum, Juan~Leon Alcazar, Merey Ramazanova, Chen Zhao, and Bernard
  Ghanem.
\newblock Just a glimpse: Rethinking temporal information for video continual
  learning.
\newblock In {\em Proceedings of the IEEE/CVF Conference on Computer Vision and
  Pattern Recognition (CVPR) Workshop}, 2023.

\bibitem{adapter_former}
Shoufa Chen, Chongjian Ge, Zhan Tong, Jiangliu Wang, Yibing Song, Jue Wang, and
  Ping Luo.
\newblock Adaptformer: Adapting vision transformers for scalable visual
  recognition.
\newblock {\em CoRR}, 2022.

\bibitem{bert}
Jacob Devlin, Ming{-}Wei Chang, Kenton Lee, and Kristina Toutanova.
\newblock {BERT:} pre-training of deep bidirectional transformers for language
  understanding.
\newblock In {\em Proceedings of the Annual Conference of the North American
  Chapter of the Association for Computational Linguistics (NAACL)}, 2019.

\bibitem{vit}
Alexey Dosovitskiy, Lucas Beyer, Alexander Kolesnikov, Dirk Weissenborn,
  Xiaohua Zhai, Thomas Unterthiner, Mostafa Dehghani, Matthias Minderer, Georg
  Heigold, Sylvain Gelly, Jakob Uszkoreit, and Neil Houlsby.
\newblock An image is worth 16x16 words: Transformers for image recognition at
  scale.
\newblock In {\em Proceedings of the International Conference on Learning
  Representations (ICLR)}, 2021.

\bibitem{podnet}
Arthur Douillard, Matthieu Cord, Charles Ollion, Thomas Robert, and Eduardo
  Valle.
\newblock Podnet: Pooled outputs distillation for small-tasks incremental
  learning.
\newblock In {\em Proceedings of the European Conference on Computer Vision
  (ECCV)}, 2020.

\bibitem{adaptbias}
Chin{-}Lun Fu, Zih{-}Ching Chen, Yun{-}Ru Lee, and Hung{-}yi Lee.
\newblock Adapterbias: Parameter-efficient token-dependent representation shift
  for adapters in {NLP} tasks.
\newblock In {\em Proceedings of the Annual Conference of the North American
  Chapter of the Association for Computational Linguistics (NAACL)}, 2022.

\bibitem{r-dfcil}
Qiankun Gao, Chen Zhao, Bernard Ghanem, and Jian Zhang.
\newblock {R-DFCIL:} relation-guided representation learning for data-free
  class incremental learning.
\newblock In {\em Proceedings of the European Conference on Computer Vision
  (ECCV)}, 2022.

\bibitem{catastrophic_forgetting}
Ian~J Goodfellow, Mehdi Mirza, Da Xiao, Aaron Courville, and Yoshua Bengio.
\newblock An empirical investigation of catastrophic forgetting in
  gradient-based neural networks.
\newblock {\em arXiv preprint arXiv:1312.6211}, 2013.

\bibitem{uniview_pet}
Junxian He, Chunting Zhou, Xuezhe Ma, Taylor Berg{-}Kirkpatrick, and Graham
  Neubig.
\newblock Towards a unified view of parameter-efficient transfer learning.
\newblock In {\em Proceedings of the International Conference on Learning
  Representations (ICLR)}, 2022.

\bibitem{mae}
Kaiming He, Xinlei Chen, Saining Xie, Yanghao Li, Piotr Doll{\'{a}}r, and
  Ross~B. Girshick.
\newblock Masked autoencoders are scalable vision learners.
\newblock In {\em Proceedings of the IEEE/CVF Conference on Computer Vision and
  Pattern Recognition (CVPR)}, 2022.

\bibitem{moco}
Kaiming He, Haoqi Fan, Yuxin Wu, Saining Xie, and Ross~B. Girshick.
\newblock Momentum contrast for unsupervised visual representation learning.
\newblock In {\em Proceedings of the IEEE/CVF Conference on Computer Vision and
  Pattern Recognition (CVPR)}, 2020.

\bibitem{resnet}
Kaiming He, Xiangyu Zhang, Shaoqing Ren, and Jian Sun.
\newblock Deep residual learning for image recognition.
\newblock In {\em Proceedings of the IEEE/CVF Conference on Computer Vision and
  Pattern Recognition (CVPR)}, 2016.

\bibitem{imagenet-r}
Dan Hendrycks, Steven Basart, Norman Mu, Saurav Kadavath, Frank Wang, Evan
  Dorundo, Rahul Desai, Tyler Zhu, Samyak Parajuli, Mike Guo, Dawn Song, Jacob
  Steinhardt, and Justin Gilmer.
\newblock The many faces of robustness: {A} critical analysis of
  out-of-distribution generalization.
\newblock In {\em Proceedings of the IEEE/CVF International Conference on
  Computer Vision (ICCV)}, 2021.

\bibitem{ucir}
Saihui Hou, Xinyu Pan, Chen~Change Loy, Zilei Wang, and Dahua Lin.
\newblock Learning a unified classifier incrementally via rebalancing.
\newblock In {\em Proceedings of the IEEE/CVF Conference on Computer Vision and
  Pattern Recognition (CVPR)}, 2019.

\bibitem{adapter_tuning}
Neil Houlsby, Andrei Giurgiu, Stanislaw Jastrzebski, Bruna Morrone, Quentin de
  Laroussilhe, Andrea Gesmundo, Mona Attariyan, and Sylvain Gelly.
\newblock Parameter-efficient transfer learning for {NLP}.
\newblock In {\em Proceedings of the International Conference on Machine
  Learning (ICML)}, 2019.

\bibitem{lora}
Edward~J. Hu, Yelong Shen, Phillip Wallis, Zeyuan Allen{-}Zhu, Yuanzhi Li,
  Shean Wang, Lu Wang, and Weizhu Chen.
\newblock Lora: Low-rank adaptation of large language models.
\newblock In {\em Proceedings of the International Conference on Learning
  Representations (ICLR)}, 2022.

\bibitem{cpg}
Steven C.~Y. Hung, Cheng{-}Hao Tu, Cheng{-}En Wu, Chien{-}Hung Chen, Yi{-}Ming
  Chan, and Chu{-}Song Chen.
\newblock Compacting, picking and growing for unforgetting continual learning.
\newblock In {\em Proceedings of the Advances in Neural Information Processing
  Systems (NeurIPS)}, 2019.

\bibitem{vpt}
Menglin Jia, Luming Tang, Bor{-}Chun Chen, Claire Cardie, Serge~J. Belongie,
  Bharath Hariharan, and Ser{-}Nam Lim.
\newblock Visual prompt tuning.
\newblock In {\em Proceedings of the European Conference on Computer Vision
  (ECCV)}, 2022.

\bibitem{ewc}
James Kirkpatrick, Razvan Pascanu, Neil Rabinowitz, Joel Veness, Guillaume
  Desjardins, Andrei~A. Rusu, Kieran Milan, John Quan, Tiago Ramalho, Agnieszka
  Grabska-Barwinska, Demis Hassabis, Claudia Clopath, Dharshan Kumaran, and
  Raia Hadsell.
\newblock Overcoming catastrophic forgetting in neural networks.
\newblock {\em Proceedings of the National Academy of Sciences (PNAS)}, 2017.

\bibitem{cifar}
Alex Krizhevsky, Geoffrey Hinton, et~al.
\newblock Learning multiple layers of features from tiny images.
\newblock {\em Technical Report}, 2009.

\bibitem{lpft}
Ananya Kumar, Aditi Raghunathan, Robbie~Matthew Jones, Tengyu Ma, and Percy
  Liang.
\newblock Fine-tuning can distort pretrained features and underperform
  out-of-distribution.
\newblock In {\em Proceedings of the International Conference on Learning
  Representations (ICLR)}. OpenReview.net, 2022.

\bibitem{cls_theory_updated}
Dharshan Kumaran, Demis Hassabis, and James~L McClelland.
\newblock What learning systems do intelligent agents need? complementary
  learning systems theory updated.
\newblock {\em Trends in cognitive sciences}, 2016.

\bibitem{ebm_tutorial}
Yann LeCun, Sumit Chopra, Raia Hadsell, M Ranzato, and Fujie Huang.
\newblock A tutorial on energy-based learning.
\newblock {\em Predicting structured data}, 2006.

\bibitem{prompt_tuning}
Brian Lester, Rami Al{-}Rfou, and Noah Constant.
\newblock The power of scale for parameter-efficient prompt tuning.
\newblock In {\em Proceedings of the Conference on Empirical Methods in Natural
  Language Processing (EMNLP)}, 2021.

\bibitem{l2g}
Xilai Li, Yingbo Zhou, Tianfu Wu, Richard Socher, and Caiming Xiong.
\newblock Learn to grow: {A} continual structure learning framework for
  overcoming catastrophic forgetting.
\newblock In {\em Proceedings of the International Conference on Machine
  Learning (ICML)}, 2019.

\bibitem{prefix_tuning}
Xiang~Lisa Li and Percy Liang.
\newblock Prefix-tuning: Optimizing continuous prompts for generation.
\newblock In {\em Proceedings of the Joint Conference of the 59th Annual
  Meeting of the Association for Computational Linguistics and the 11th
  International Joint Conference on Natural Language Processing (ACL-IJCNLP
  2021)}, 2021.

\bibitem{lwf}
Zhizhong Li and Derek Hoiem.
\newblock Learning without forgetting.
\newblock {\em IEEE Transactions on Pattern Analysis and Machine Intelligence
  (TPAMI)}, 2017.

\bibitem{eb_ood}
Weitang Liu, Xiaoyun Wang, John~D. Owens, and Yixuan Li.
\newblock Energy-based out-of-distribution detection.
\newblock In {\em Proceedings of the Advances in Neural Information Processing
  Systems (NeurIPS)}, 2020.

\bibitem{aanets}
Yaoyao Liu, Bernt Schiele, and Qianru Sun.
\newblock Adaptive aggregation networks for class-incremental learning.
\newblock In {\em Proceedings of the IEEE/CVF Conference on Computer Vision and
  Pattern Recognition (CVPR)}, 2021.

\bibitem{swin}
Ze Liu, Yutong Lin, Yue Cao, Han Hu, Yixuan Wei, Zheng Zhang, Stephen Lin, and
  Baining Guo.
\newblock Swin transformer: Hierarchical vision transformer using shifted
  windows.
\newblock In {\em Proceedings of the IEEE/CVF International Conference on
  Computer Vision (ICCV)}, 2021.

\bibitem{convnext}
Zhuang Liu, Hanzi Mao, Chao{-}Yuan Wu, Christoph Feichtenhofer, Trevor Darrell,
  and Saining Xie.
\newblock A convnet for the 2020s.
\newblock In {\em Proceedings of the IEEE/CVF Conference on Computer Vision and
  Pattern Recognition (CVPR)}, 2022.

\bibitem{compacter}
Rabeeh~Karimi Mahabadi, James Henderson, and Sebastian Ruder.
\newblock Compacter: Efficient low-rank hypercomplex adapter layers.
\newblock In {\em Proceedings of the Advances in Neural Information Processing
  Systems (NeurIPS)}, 2021.

\bibitem{cls_theory}
James~L McClelland, Bruce~L McNaughton, and Randall~C O'Reilly.
\newblock Why there are complementary learning systems in the hippocampus and
  neocortex: insights from the successes and failures of connectionist models
  of learning and memory.
\newblock {\em Psychological review}, 1995.

\bibitem{domainnet}
Xingchao Peng, Qinxun Bai, Xide Xia, Zijun Huang, Kate Saenko, and Bo Wang.
\newblock Moment matching for multi-source domain adaptation.
\newblock In {\em Proceedings of the IEEE/CVF International Conference on
  Computer Vision (ICCV)}, pages 1406--1415, 2019.

\bibitem{gpt}
Alec Radford, Karthik Narasimhan, Tim Salimans, Ilya Sutskever, et~al.
\newblock Improving language understanding by generative pre-training.
\newblock 2018.

\bibitem{res_adapter}
Sylvestre{-}Alvise Rebuffi, Hakan Bilen, and Andrea Vedaldi.
\newblock Learning multiple visual domains with residual adapters.
\newblock In {\em Proceedings of the Advances in Neural Information Processing
  Systems (NeurIPS)}, 2017.

\bibitem{icarl}
Sylvestre-Alvise Rebuffi, Alexander Kolesnikov, Georg Sperl, and Christoph~H
  Lampert.
\newblock icarl: Incremental classifier and representation learning.
\newblock In {\em Proceedings of the IEEE/CVF Conference on Computer Vision and
  Pattern Recognition (CVPR)}, 2017.

\bibitem{cfrp}
Anthony~V. Robins.
\newblock Catastrophic forgetting, rehearsal and pseudorehearsal.
\newblock {\em Connect. Sci.}, 1995.

\bibitem{imagenet}
Olga Russakovsky, Jia Deng, Hao Su, Jonathan Krause, Sanjeev Satheesh, Sean Ma,
  Zhiheng Huang, Andrej Karpathy, Aditya Khosla, Michael Bernstein, et~al.
\newblock Imagenet large scale visual recognition challenge.
\newblock {\em International Journal of Computer Vision (IJCV)}, 2015.

\bibitem{pnn}
Andrei~A. Rusu, Neil~C. Rabinowitz, Guillaume Desjardins, Hubert Soyer, James
  Kirkpatrick, Koray Kavukcuoglu, Razvan Pascanu, and Raia Hadsell.
\newblock Progressive neural networks.
\newblock {\em CoRR}, 2016.

\bibitem{dgr}
Hanul Shin, Jung~Kwon Lee, Jaehong Kim, and Jiwon Kim.
\newblock Continual learning with deep generative replay.
\newblock In {\em Proceedings of the Advances in Neural Information Processing
  Systems (NeurIPS)}, 2017.

\bibitem{abd}
James Smith, Yen-Chang Hsu, Jonathan Balloch, Yilin Shen, Hongxia Jin, and
  Zsolt Kira.
\newblock Always be dreaming: A new approach for data-free class-incremental
  learning.
\newblock In {\em Proceedings of the IEEE/CVF International Conference on
  Computer Vision (ICCV)}, 2021.

\bibitem{coda_prompt}
James~Seale Smith, Leonid Karlinsky, Vyshnavi Gutta, Paola Cascante-Bonilla,
  Donghyun Kim, Assaf Arbelle, Rameswar Panda, Rogerio Feris, and Zsolt Kira.
\newblock Coda-prompt: Continual decomposed attention-based prompting for
  rehearsal-free continual learning.
\newblock In {\em Proceedings of the IEEE/CVF Conference on Computer Vision and
  Pattern Recognition (CVPR)}, 2023.

\bibitem{esn}
Yabin Wang, Zhiheng Ma, Zhiwu Huang, Yaowei Wang, Zhou Su, and Xiaopeng Hong.
\newblock Isolation and impartial aggregation: A paradigm of incremental
  learning without interference.
\newblock 2023.

\bibitem{dual_prompt}
Zifeng Wang, Zizhao Zhang, Sayna Ebrahimi, Ruoxi Sun, Han Zhang, Chen{-}Yu Lee,
  Xiaoqi Ren, Guolong Su, Vincent Perot, Jennifer~G. Dy, and Tomas Pfister.
\newblock Dualprompt: Complementary prompting for rehearsal-free continual
  learning.
\newblock In {\em Proceedings of the European Conference on Computer Vision
  (ECCV)}, 2022.

\bibitem{l2p}
Zifeng Wang, Zizhao Zhang, Chen{-}Yu Lee, Han Zhang, Ruoxi Sun, Xiaoqi Ren,
  Guolong Su, Vincent Perot, Jennifer~G. Dy, and Tomas Pfister.
\newblock Learning to prompt for continual learning.
\newblock In {\em Proceedings of the IEEE/CVF Conference on Computer Vision and
  Pattern Recognition (CVPR)}, 2022.

\bibitem{ufo}
Teng Xi, Yifan Sun, Deli Yu, Bi Li, Nan Peng, Gang Zhang, Xinyu Zhang, Zhigang
  Wang, Jinwen Chen, Jian Wang, Lufei Liu, Haocheng Feng, Junyu Han, Jingtuo
  Liu, Errui Ding, and Jingdong Wang.
\newblock {UFO:} unified feature optimization.
\newblock In {\em Proceedings of the European Conference on Computer Vision
  (ECCV)}, 2022.

\bibitem{deepinversion}
Hongxu Yin, Pavlo Molchanov, Jose~M. Alvarez, Zhizhong Li, Arun Mallya, Derek
  Hoiem, Niraj~K. Jha, and Jan Kautz.
\newblock Dreaming to distill: Data-free knowledge transfer via deepinversion.
\newblock In {\em Proceedings of the IEEE/CVF Conference on Computer Vision and
  Pattern Recognition (CVPR)}, 2020.

\bibitem{den}
Jaehong Yoon, Eunho Yang, Jeongtae Lee, and Sung~Ju Hwang.
\newblock Lifelong learning with dynamically expandable networks.
\newblock In {\em Proceedings of the International Conference on Learning
  Representations (ICLR)}, 2018.

\bibitem{bitfit}
Elad~Ben Zaken, Yoav Goldberg, and Shauli Ravfogel.
\newblock Bitfit: Simple parameter-efficient fine-tuning for transformer-based
  masked language-models.
\newblock In {\em Proceedings of the Annual Meeting of the Association for
  Computational Linguistics (ACL)}, 2022.

\bibitem{pi}
Friedemann Zenke, Ben Poole, and Surya Ganguli.
\newblock Continual learning through synaptic intelligence.
\newblock In {\em Proceedings of the International Conference on Machine
  Learning (ICML)}, 2017.

\bibitem{side_tuning}
Jeffrey~O. Zhang, Alexander Sax, Amir Zamir, Leonidas~J. Guibas, and Jitendra
  Malik.
\newblock Side-tuning: {A} baseline for network adaptation via additive side
  networks.
\newblock In {\em Proceedings of the European Conference on Computer Vision
  (ECCV)}, 2020.

\end{thebibliography}
}

\appendix
\clearpage

\renewcommand{\thetable}{\Roman{table}}
\renewcommand{\thefigure}{\Roman{figure}}
\renewcommand{\theequation}{\Roman{equation}}
\renewcommand{\thesection}{\Alph{section}}

\twocolumn[
  \begin{@twocolumnfalse}
    \begin{center}
      \Large\textbf{Supplementary Materials}
      \vspace{20pt}
    \end{center}
  \end{@twocolumnfalse}
]

In the supplementary materials, we further validate the proposed LAE framework by providing the following:
\begin{itemize}
  \item Section~\ref{sec:exp_details}: Additional Experimental Details.
  \item Section~\ref{sec:exp_results}: Additional Experimental Results.
  \item Section~\ref{sec:prompt_pool}: Investigation on Prompt Learning and Selection from Pool in Prompt-Pool-based Approaches.
\end{itemize}

\section{Additional Experimental Details}
\label{sec:exp_details}
\noindent\textbf{Data Augmentation}. We adopt a very simple data augmentation strategy for training, following L2P~\cite{l2p} and DualPrompt~\cite{dual_prompt}.
\emph{1)} Images are randomly resized to $224\times224$ using the bilinear interpolation algorithm. \emph{2)} Images are normalized by min-max (for ViT~\cite{vit}) or standard deviation (for Swin Transformer~\cite{swin} and ConvNeXt~\cite{convnext}) normalization. \emph{3)} Images are randomly flipped from horizontal. During inference, images are resized to $256\times256$ and cropped to $224\times224$ from central. All other approaches take the same data augmentation strategy as ours for fair comparisons.
The PyTorch-like code is present in Algorithm~\ref{alg:data_aug}.

\noindent\textbf{Hyper-Parameter}. Our LAE introduced  two additional hyper-parameters, \ie, the weight decay $\alpha$ of the Exponential Moving Average (EMA) algorithm and freezing epochs of the online Parameter-Efficient Tuning (PET) module. We did not intentionally search for these parameters and set $\alpha$ to a value very close to 1, such as the default value $0.9999$ we used. The number of freezing epochs can be determined by the change in loss after freezing the online PET module and is typically set to the value where the loss no longer decreases. We set this value to 3 for CIFAR100 and scaled it proportionally for ImageNet-R, on all of which our LAE achieved superior performance than other competitors.

\begin{algorithm*}[t]
  \caption{{Data Augmentation Code (PyTorch-like)}}
  \label{alg:data_aug}
  \definecolor{codeblue}{rgb}{0.25,0.5,0.25}
  \lstset{
    backgroundcolor=\color{white},
    basicstyle=\fontsize{7.2pt}{7.2pt}\ttfamily\selectfont,
    columns=fullflexible,
    breaklines=true,
    captionpos=b,
    commentstyle=\fontsize{7.2pt}{7.2pt}\color{codeblue},
    keywordstyle=\fontsize{7.2pt}{7.2pt},
  }
  \begin{lstlisting}[language=python]
def build_train_transform(model):
    transforms = [T.RandomResizedCrop(224), T.RandomHorizontalFlip(), T.ToTensor()]
    if not isinstance(model, VisionTransformer):
        transforms.append(T.Normalize(mean=IMAGENET_MEAN, std=IMAGENET_STD))
    return T.Compose(transforms)

def build_inference_transform(model):
    transforms = [T.Resize(256), T.CenterCrop(224), T.ToTensor()]
    if not isinstance(model, VisionTransformer):
        transforms.append(T.Normalize(mean=IMAGENET_MEAN, std=IMAGENET_STD))
    return T.Compose(transforms)
\end{lstlisting}
\end{algorithm*}

\begin{algorithm*}[t]
  \caption{{Training and Inference Code (PyTorch-like)}}
  \label{alg:lae_train}
  \definecolor{codeblue}{rgb}{0.25,0.5,0.25}
  \lstset{
    backgroundcolor=\color{white},
    basicstyle=\fontsize{7.2pt}{7.2pt}\ttfamily\selectfont,
    columns=fullflexible,
    breaklines=true,
    captionpos=b,
    commentstyle=\fontsize{7.2pt}{7.2pt}\color{codeblue},
    keywordstyle=\fontsize{7.2pt}{7.2pt},
  }

  \begin{lstlisting}[language=python]
# model: the pre-trained model; pet_on: online PET module; pet_off: offline PET module;

def train(model, pet_on, pet_off, dataloader, optimizer, task_id, alpha):
    model = attach(model, pet_on)
    for e in range(MAX_EPOCHS):
        if e == 0 and not_the_first_task(task_id):
            freeze(pet_on)
        elif e == NUM_FREEZING_EPOCHS:
            unfreeze(pet_on)
        
        for input, target in dataloader:
            pred = mask(model(input), task_id) # Eq. (8) in the paper
            loss = cross_entropy(pred, target) # Eq. (8) in the paper
            optimizer.zero_grad()
            loss.backward()
            optimizer.step()
            ema_update(pet_off, pet_on, alpha) # Eq. (13) in the paper

def inference(model, pet_on, pet_off, input):
    pred_on, pred_off = attach(model, pet_on)(input), attach(model, pet_off)(input) 
    pred = max(softmax(pred_on, dim=-1), softmax(pref_off, dim=-1)) # Eq. (14) in the paper
    return argmax(pred)
\end{lstlisting}
\end{algorithm*}

\noindent\textbf{Training, Inference and Evaluation}. The training and inference of our LAE framework are very easy to implement, the PyTorch-like pseudocode is provided in Algorithms~\ref{alg:lae_train}.  It is important to note that our evaluation metric $A_{10}$ (Equation 15 in the paper) is slightly different from the following metric used by original L2P and DualPrompt:
\begin{equation}
  A_{10} = \frac{1}{10} \sum_{j=1}^{10}
  \frac{1}{\lvert \mathcal{D}_{j}^{test} \rvert}
  \sum_{(\mathbf{x}, y) \in \mathcal{D}_{j}^{test}}
  \mathds{1} \left(\hat{y} = y\right),
\end{equation}
where $\mathcal{D}_{j}^{test}$ is the test set of the $j^{th}$ task. We train and evaluate on three different class orders, while L2P, DualPrompt, and ESN~\cite{esn} only evaluate on one class order in their original papers. Additionally, ESN uses a different pre-trained checkpoint from L2P and DualPrompt, but we correct this issue when using its code. The above differences lead to slightly different experimental results reported in their original papers from the data reported by us.

\begin{table}
  \renewcommand\arraystretch{1.0}
  \caption{20-Task Benchmark Results on CIFAR100. The PET modules are inserted into the first 5 transformer blocks of the standard ViT-B/16 pre-trained on the ImageNet21k dataset. The ``5, 10, 20'' indicate the size of PET modules.}
  \vskip -0.25in
  \label{table:cifar100_t20}
  \begin{center}
    \resizebox{\columnwidth}{!}{
      \begin{tabular}{lcccr}
        \toprule
        Approach                      & PET Module & $A_{20}$ (↑)            & $\bar{A}_{20}$ (↑)      \\
        \midrule
        L2P~\cite{l2p}                & Prompt     & 80.10$\pm$0.72          & 85.29$\pm$0.50          \\
        DualPrompt~\cite{dual_prompt} & Prefix20   & 82.02$\pm$0.32          & 89.50$\pm$0.11          \\
        ESN~\cite{esn}                & Prompt     & 80.56$\pm$0.94          & 90.47$\pm$1.19          \\
        \midrule
        \multirow{6}{*}{LAE (Ours)}   & Adapter5   & 83.89$\pm$0.60          & \textbf{92.35}$\pm$0.55 \\
                                      & Adapter10  & 83.81$\pm$0.35          & 92.32$\pm$0.57          \\
                                      & LoRA5      & 83.92$\pm$0.36          & 92.15$\pm$0.47          \\
                                      & LoRA10     & 83.35$\pm$0.20          & 91.71$\pm$0.88          \\
                                      & Prefix10   & 83.82$\pm$0.18          & 92.07$\pm$0.72          \\
                                      & Prefix20   & \textbf{83.93}$\pm$0.28 & 92.21$\pm$0.53          \\
        \bottomrule
      \end{tabular}
    }
  \end{center}
  \vskip -0.2in
\end{table}

\begin{table}
  \renewcommand\arraystretch{1.0}
  \caption{20-Task Benchmark Results on ImageNet-R. The PET modules are inserted into the first 5 transformer blocks of the standard ViT-B/16 pre-trained on the ImageNet21k dataset. The ``5, 10, 20'' indicate the size of PET modules.}
  \vskip -0.25in
  \label{table:imagenet-r_t20}
  \begin{center}
    \resizebox{\columnwidth}{!}{
      \begin{tabular}{lcccr}
        \toprule
        Approach                      & PET Module & $A_{20}$ (↑)            & $\bar{A}_{20}$ (↑)      \\
        \midrule
        L2P~\cite{l2p}                & Prompt     & 59.85$\pm$1.38          & 66.33$\pm$2.46          \\
        DualPrompt~\cite{dual_prompt} & Prefix20   & 66.61$\pm$0.24          & 76.94$\pm$1.39          \\
        ESN~\cite{esn}                & Prompt     & 58.65$\pm$0.83          & 70.94$\pm$1.88          \\
        \midrule
        \multirow{6}{*}{LAE (Ours)}   & Adapter5   & 69.66$\pm$1.16          & 81.69$\pm$1.00          \\
                                      & Adapter10  & 69.19$\pm$1.25          & \textbf{81.78}$\pm$0.77 \\
                                      & LoRA5      & 68.91$\pm$1.40          & 80.99$\pm$1.17          \\
                                      & LoRA10     & 69.07$\pm$1.49          & 81.12$\pm$1.09          \\
                                      & Prefix10   & \textbf{69.67}$\pm$0.86 & 79.97$\pm$0.97          \\
                                      & Prefix20   & 69.34$\pm$0.84          & 79.90$\pm$1.08          \\
        \bottomrule
      \end{tabular}
    }
  \end{center}
  \vskip -0.4in
\end{table}

\section{Additional Experimental Results}
\label{sec:exp_results}
\begin{figure*}[t]
  \begin{center}
    \begin{subfigure}[b]{0.49\textwidth}
      \centering
      \includegraphics[width=\textwidth]{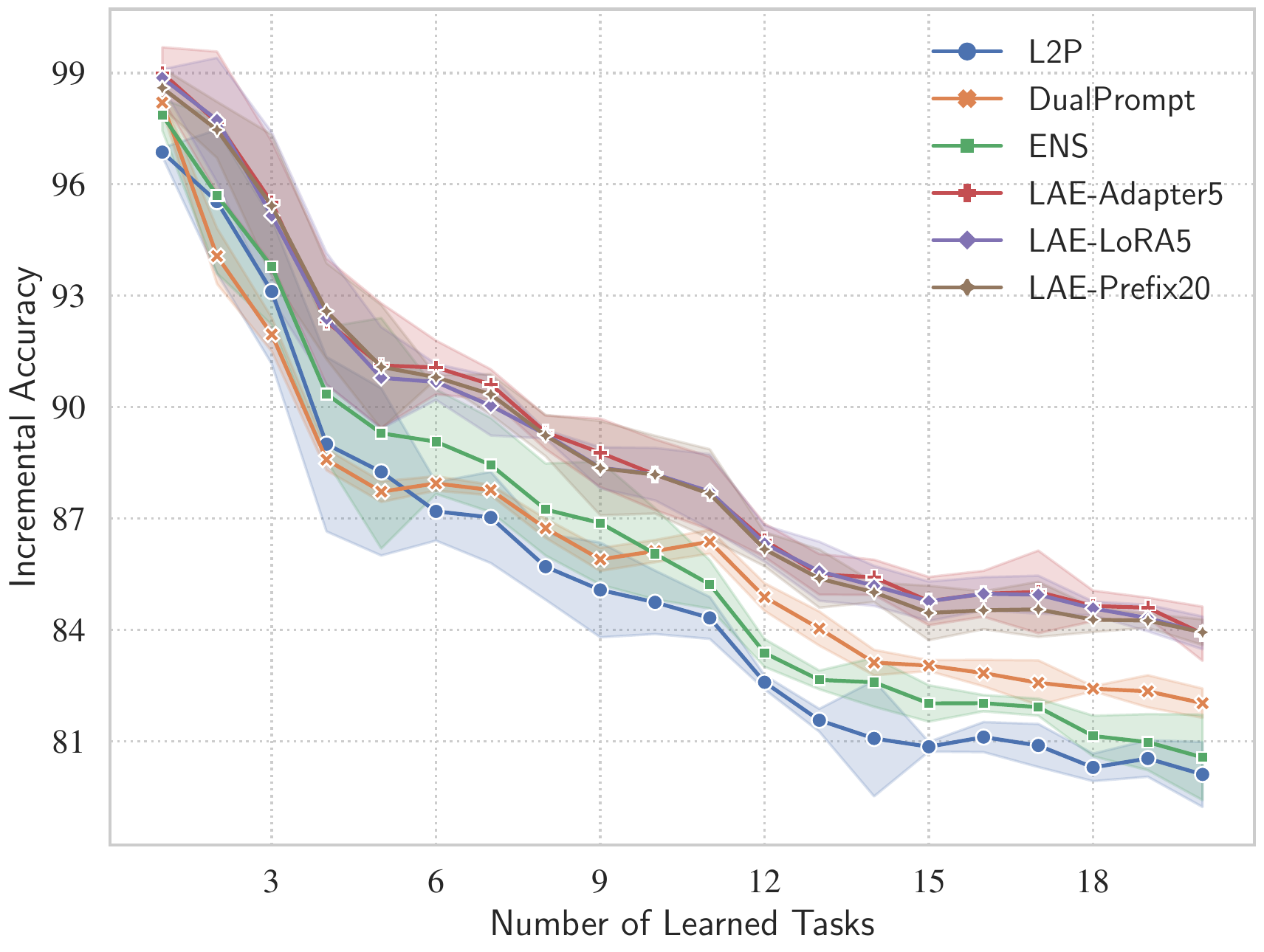}
      \caption{CIFAR100}
      \label{fig:benchmark_cifar_t20}
    \end{subfigure}
    \hfill
    \begin{subfigure}[b]{0.49\textwidth}
      \centering
      \includegraphics[width=\textwidth]{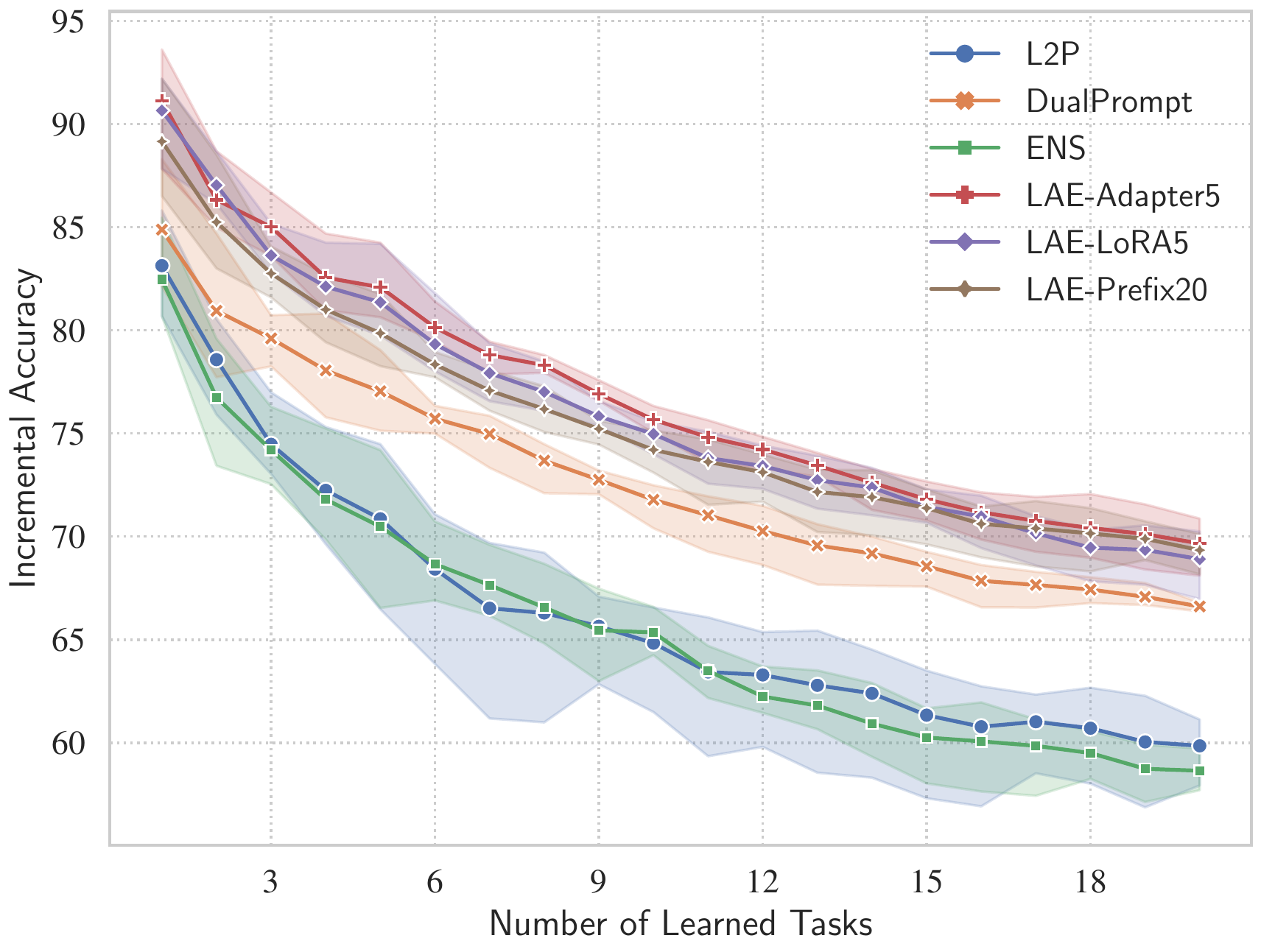}
      \caption{ImageNet-R}
      \label{fig:benchmark_imgr_t20}
    \end{subfigure}
    \hfill
    \vskip -0.1in
    \caption{Task-by-Task Incremental Accuracy on two 20-task benchmarks. The lines illustrate the task-by-task evaluation results of L2P~\cite{l2p}, DualPrompt~\cite{dual_prompt}, ENS~\cite{esn}, and our LAE framework with different PET modules.}
    \label{fig:benchmarks_t20}
  \end{center}
  \vskip -0.3in
\end{figure*}

\noindent\textbf{20-Task Benchmark Results}.
To further validate the efficacy of our LAE in longer-term Continual Learning scenarios, we split the CIFAR100~\cite{cifar} and ImageNet-R~\cite{imagenet} datasets into 20 tasks, each containing 5 (for CIFAR100) or 10 (for ImageNet-R) classes. We then conducted experiments and reported the mean and standard deviation of three runs in different class orders in Tables~\ref{table:cifar100_t20} and~\ref{table:imagenet-r_t20}.
Similar to the 10-task experiments in the paper, we plot the task-by-task evaluation results in Figure~\ref{fig:benchmark_cifar_t20} and~\ref{fig:benchmark_imgr_t20} for the 20-task experiments on CIFAR100 and ImageNet-R. From these tables and figures, we can observe a wider performance gap between our LAE and other competitors compared to the 10-task experiments, suggesting that our LAE is more effective at mitigating forgetting and achieving a better stability-plasticity balance in longer-term Continual Learning.

\begin{table}
  \renewcommand\arraystretch{1.0}
  \caption{Comparison with CODA-Prompt on 5-task DomainNet Benchmark, 5- and 10-task ImageNet-R benchmarks. ``$A_N$'' and ``$F_N$'' are last incremental accuracy and last average forgetting for N-task benchmarks respectively.}
  \vskip -0.25in
  \label{table:coda_comparasion}
  \begin{center}
    \resizebox{\columnwidth}{!}{
      \begin{tabular}{lccccc}
        \toprule
        \multicolumn{3}{c}{Approach} & Joint-FT                   & CODA-Prompt  & LAE (Prefix10)                 \\
        \midrule
        \multirow{2.5}{*}{DomainNet} & \multirow{2.5}{*}{5-task}  & $A_5$ (↑)    & 74.91          & 67.11 & 68.37 \\
        \cmidrule{3-6}
                                     &                            & $F_5$ (↓)    & -              & 13.79 & 8.33  \\
        \midrule
        \multirow{5}{*}{ImageNet-R}  & \multirow{2.5}{*}{5-task}  & $A_{5}$ (↑)  & 81.08          & 75.32 & 76.69 \\
        \cmidrule{3-6}
                                     &                            & $F_{5}$ (↓)  & -              & 6.09  & 6.17  \\
        \cmidrule{2-6}
                                     & \multirow{2.5}{*}{10-task} & $A_{10}$ (↑) & 81.08          & 74.31 & 74.43 \\
        \cmidrule{3-6}
                                     &                            & $F_{10}$ (↓) & -              & 5.63  & 5.22  \\
        \bottomrule
      \end{tabular}
    }
  \end{center}
  \vskip -0.2in
\end{table}

\smallskip
\noindent\textbf{Comparison with CODA-Prompt}.
The contemporary CODA-Prompt~\cite{coda_prompt} approach demonstrates remarkable performance. Nevertheless, upon reviewing the authors' released code, we identified three potential sources of unfair comparison:
1) A distinct ImageNet-R train-test split in contrast to DualPrompt.
2) The model is pretrained on ImageNet-21k and subsequently fine-tuned on ImageNet-1K.
3) Varied training strategies, such as the number of epochs and learning rates.
Our initial experiments reveal that when utilizing DualPrompt's train-test split, CODA-Prompt consistently underperforms our LAE. To ensure a fair evaluation, we adopt CODA-Prompt's settings for our experiments and extend our assessment to the DomainNet~\cite{domainnet} dataset. All results are showcased in Table~\ref{table:coda_comparasion}, where we present average forgetting rates instead of average incremental accuracy.

\begin{table}
  \renewcommand\arraystretch{1.0}
  \caption{The sensitiveness w.r.t. EMA's weight decay $\alpha$.}
  \vskip -0.25in
  \label{table:ablation_alpha}
  \begin{center}
    \resizebox{\columnwidth}{!}{
      \begin{tabular}{ccccr}
        \toprule
        $\alpha$           & 0.999          & 0.9999         & 0.99999        \\
        \midrule
        $A_{10}$ (↑)       & 71.40$\pm$1.02 & 72.66$\pm$0.63 & 72.58$\pm$0.40 \\
        $\bar{A}_{10}$ (↑) & 78.04$\pm$1.03 & 78.91$\pm$0.89 & 78.67$\pm$0.94 \\
        \bottomrule
      \end{tabular}
    }
  \end{center}
  \vskip -0.3in
\end{table}

\smallskip
\noindent\textbf{Sensitive Analysis on EMA's Weight Decay}.
Weight decay $\alpha$ plays an important role in the knowledge accumulation of the offline PET module. A small value can lead to the integration of too much unstable new knowledge during the learning process, while a large value can result in the offline PET module being unable to effectively absorb new knowledge. In all of our experiments in the paper, we set the weight decay of EMA to 0.9999, which is the default value in the timm\footnote{https://github.com/huggingface/pytorch-image-models} library. Our experimental results in Table~\ref{table:ablation_alpha} demonstrate that this value yields the best performance.

\smallskip
\noindent\textbf{Memory and computation complexity}.
Our LAE requires two forward passes (one with $\boldsymbol{\theta}{pet}^{off}$ and the other with $\boldsymbol{\theta}{pet}^{on}$) per inference sample, yielding computational costs on par with L2P, DualPrompt, and the contemporary approach CODA-Prompt. Additionally, due to the constant number of parameters maintained across all tasks, LAE introduces fewer new parameters, as illustrated in Table~\ref{tab:param_com}.

\begin{table}
  \renewcommand\arraystretch{1.0}
  \caption{The statistics of introduced parameters by approaches on 10-task benchmarks. ``A10'', ``L10'' and ``P20'' indicate Adapter10, LoRA10 and Prefix20 respectively.}
  \vskip -0.25in
  \label{tab:param_com}
  \begin{center}
    \resizebox{\columnwidth}{!}{
      \begin{tabular}{ccccc}
        \toprule
        Approach     & DualPrompt & LAE (A10) & LAE (L10) & LAE (P20) \\
        \midrule
        \#Param. (M) & 1.03       & 0.15      & 0.29      & 0.29      \\
        \bottomrule
      \end{tabular}
    }
  \end{center}
  \vskip -0.2in
\end{table}

\begin{table*}
  \renewcommand\arraystretch{0.85}
  \caption{Evaluation results on all tasks using 10 sets of task-specific E-Prompts. ``\#E-Prompts'' denotes the index of the E-Prompts, \eg, ``1'' indicates evaluation using the first task's E-Prompts.}
  \vskip -0.2in
  \label{table:dp_metrics}
  \begin{center}
    \resizebox{\textwidth}{!}{
      \begin{tabular}{lccccccccccr}
        \toprule
        \#E-Prompts        & 1     & 2     & 3     & 4     & 5     & 6     & 7     & 8     & 9     & 10    \\
        \midrule
        $A_{10}$ (↑)       & 63.78 & 66.93 & 67.57 & 67.88 & 68.48 & 68.38 & 68.80 & 68.88 & 68.87 & 68.47 \\
        $\bar{A}_{10}$ (↑) & 69.11 & 72.54 & 72.60 & 72.63 & 72.69 & 72.68 & 72.72 & 72.73 & 72.73 & 72.69 \\
        \bottomrule
      \end{tabular}
    }
  \end{center}
  \vskip -0.2in
\end{table*}

\begin{table*}
  \renewcommand\arraystretch{0.85}
  \caption{Evaluation results on each task using 10 sets of task-specific E-Prompts. ``\#E-Prompts'' denotes the index of the E-Prompts, \eg, ``1'' indicates evaluation using the first task's E-Prompts.}
  \vskip -0.2in
  \label{table:dp_10tasks}
  \begin{center}
    \resizebox{\textwidth}{!}{
      \begin{tabular}{lccccccccccr}
        \toprule
        \#E-Prompts & 1     & 2     & 3     & 4     & 5     & 6     & 7     & 8     & 9     & 10    \\
        \midrule
        Task 1      & 69.64 & 73.27 & 72.77 & 72.94 & 73.76 & 73.10 & 72.61 & 72.94 & 72.44 & 72.28 \\
        Task 2      & 67.23 & 71.73 & 71.03 & 70.89 & 71.87 & 71.03 & 71.31 & 71.31 & 71.59 & 70.61 \\
        Task 3      & 63.93 & 67.16 & 70.90 & 71.89 & 70.40 & 69.90 & 70.65 & 70.40 & 69.40 & 69.90 \\
        Task 4      & 54.61 & 60.17 & 64.52 & 68.35 & 67.48 & 67.30 & 64.52 & 64.35 & 63.65 & 62.96 \\
        Task 5      & 61.01 & 64.52 & 65.64 & 65.36 & 68.16 & 68.16 & 67.04 & 65.50 & 65.22 & 65.78 \\
        Task 6      & 59.33 & 63.73 & 63.73 & 65.14 & 66.37 & 67.43 & 67.43 & 66.55 & 66.20 & 66.20 \\
        Task 7      & 57.02 & 58.93 & 58.41 & 57.71 & 57.89 & 57.71 & 61.18 & 62.22 & 62.39 & 60.49 \\
        Task 8      & 61.47 & 61.92 & 62.14 & 61.03 & 61.47 & 62.81 & 65.26 & 68.37 & 67.26 & 65.03 \\
        Task 9      & 74.85 & 76.50 & 74.70 & 75.15 & 75.30 & 74.70 & 77.25 & 77.10 & 78.89 & 77.40 \\
        Task 10     & 65.53 & 67.72 & 69.36 & 68.40 & 68.95 & 68.95 & 68.81 & 69.08 & 69.63 & 71.41 \\
        \bottomrule
      \end{tabular}
    }
  \end{center}
  \vskip -0.2in
\end{table*}

\section{Prompt Learning and Selection from Pool}
\label{sec:prompt_pool}
L2P~\cite{l2p} and DualPrompt~\cite{dual_prompt} are two representative approaches that leverage prompt tuning~\cite{prompt_tuning} to address the problem of Continual Learning. L2P first proposes to use a pool to store prompts shared across tasks, where a set of prompts that match the sample are selected from the pool to predict the sample's label. In contrast, DualPrompt directly learns a set of task-specific E-Prompts for each task and stores them in the pool. During inference, the best-matched prompts (\ie, the prompts learned for the task that the sample belongs to) are selected for the given sample.

The performance of these approaches is influenced by two key factors. \emph{1)} The ability to learn optimal prompts for each task is crucial for achieving better plasticity, \ie, the ability to learn new knowledge. Better performance can be achieved only by sufficiently learning new knowledge while retaining as much previous knowledge as possible. \emph{2)} the ability to accurately select the best-matched prompts for the inference sample is more critical. Because even if optimal prompts are learned for each task, inference using the wrong prompts can still result in poor prediction results. Following, we take DualPrompt as an example to investigate these two abilities of prompt-pool-based approaches.

\begin{figure}[t]
  \begin{center}
    \centerline{\includegraphics[width=\columnwidth]{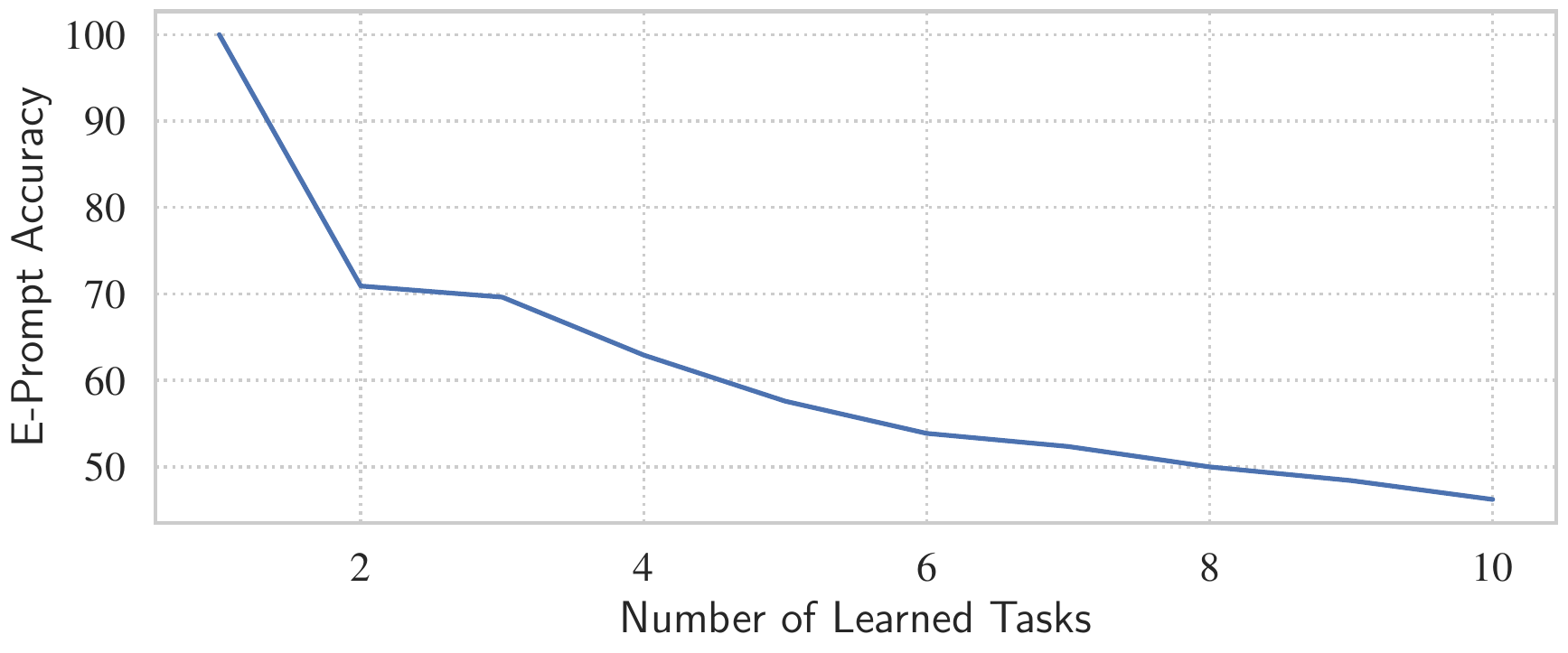}}
    \vskip -0.1in
    \caption{The E-Prompts selection accuracy of DualPrompt on the test set of ImageNet-R.}
    \label{fig:e-prompt_acc}
  \end{center}
  \vskip -0.5in
\end{figure}

To begin with, we assume that DualPrompt can learn the optimal E-prompts for each task. We then evaluate whether it can accurately select the right E-prompts during inference. As shown in Figure~\ref{fig:e-prompt_acc}, we observe that the E-prompts selected by DualPrompt are completely accurate after learning the first task. However, as the number of learned tasks increases, the accuracy of the prompt selection gradually decreases. By the time the 10th task is learned, the selection accuracy drops to below 50\%. Therefore, we conclude that if DualPrompt cannot address this issue, it is difficult to apply it to longer-term Continual Learning scenarios.

In addition, according to Figure 3 in our paper, we observe that DualPrompt performs worse than our LAE when learning the first task, regardless of whether our LAE uses Adapter~\cite{adapter_tuning} with fewer parameters, the LoRA~\cite{lora} with the equivalent number of parameters, or Prefix~\cite{prefix_tuning} (\ie, DualPrompt's E-Prompt) with slightly more parameters than DualPrompt. This indicates that Prompt/Prefix may not be as effective as Adapter and LoRA in learning new knowledge on these two datasets, as well as DualPrompt could not learn the optimal E-Prompts for the first task because they did not calibrate the Prefix like our LAE. This suggests that it is necessary to explore different Parameter-Efficient Tuning (PET) methods and calibrate PET modules.

Moreover, our naive baseline only uses one set of Prefixes, while DualPrompt learns a set of Prefixes for each task, totaling 10 sets, yet they achieve similar performance. We evaluate all 10 tasks using the 10 sets of E-Prompts learned by DualPrompt separately, and the results in Table~\ref{table:dp_metrics} show that the differences in the last and average incremental accuracy using the 2nd-10th sets of E-Prompts are very small. Table~\ref{table:dp_10tasks} presents the prediction results of each task using each set of E-Prompts, for most tasks, the prediction results using the 2nd-10th sets of E-Prompts are very close. These analyses reveal that from the learning of the second task, task-specific E-Prompts tend to become homogeneous. According to our analysis in the paper, an important reason for this is that the adaptation speed of the Prefix is much slower than classifiers and other PET modules (\ie, Adapter~\cite{adapter_tuning} and LoRA~\cite{lora}).

\end{document}